\documentclass[10pt]{article}

\pdfoutput=1 
\usepackage{amsmath}
\usepackage{amssymb}
\usepackage{graphicx}
\usepackage{cite}
\usepackage{color} 
\usepackage{dcolumn}% Align table columns on decimal point
\usepackage{bm}% bold math

\makeatletter
\renewcommand{\@biblabel}[1]{\quad#1.}
\makeatother

\date{}

\begin{document}

%\title{The effect of linguistic constraints on the large scale organization of language}
%\maketitle

%\begin{abstract}
\begin{flushleft}
{\Large
\textbf{The effect of linguistic constraints on the large scale organization of language}
}
% Insert Author names, affiliations and corresponding author email.
\\
Madhav Krishna$^{1}$, 
Ahmed Hassan$^{2}$, 
Yang Liu$^{2}$,
Dragomir Radev$^{2\ast}$
\\
\bf{1} Columbia University New York, New York, USA
\\
\bf{2} University of Michigan Ann Arbor, Michigan, USA
\\
$\ast$ E-mail: Corresponding radev@umich.edu
\end{flushleft}

% Please keep the abstract between 250 and 300 words
\section*{Abstract}

This paper studies the effect of linguistic constraints on the large scale organization of language.
It describes the properties of linguistic networks built using
texts of written language with the words randomized. These properties
are compared to those obtained for a network built over the text in
natural order. It is observed that the ``random'' networks too exhibit
small-world and scale-free characteristics. They also show a high
degree of clustering. This is indeed a surprising result - one that
has not been addressed adequately in the literature. We hypothesize
that many of the network statistics reported here studied are in fact
functions of the distribution of the underlying data from which the
network is built and may not be indicative of the nature of the
concerned network.
%\end{abstract}

\section{Introduction}

Human language is a good example of a naturally occurring
self-organizing complex system. Language, when modeled as a network,
has been shown to exhibit small-world and scale-free properties
\cite{mehler}. A network is said to be scale-free if its degree
distribution follows a a power-law \cite{barabasialbert}. Moreover, it
has also been suggested that the network evolves over time by
following the rule of preferential attachment. That is, a new word
being introduced into the system tends to associate with a
pre-existing word that is highly frequent
\cite{dorogovstevmendes}. However, probably the first ever
mathematical law that hinted at the complex nature of language was
given by Zipf \cite{zipf}. Zipf's law gives us the following
relationship between the frequency of a word, $F(r)$, in a language
and its rank $r$, when words are ranked in order of their frequencies
of occurrence (most frequent word assigned rank 1):
\begin{equation}
F(r)=\frac{C}{r^\alpha}
\end{equation}
Zipf observed this ``law'' for English (with $C\approx 0.1$ and
$\alpha \approx 1$) \cite{miller}, \cite{millerchomsky} but it has
been shown to be obeyed by other natural
languages\cite{Gelbukh&Sidorov.01}. It seems surprising at first that
Zipf's law is also obeyed by random texts \cite{miller},
\cite{millerchomsky}. Random texts may be produced artificially,
composed of symbols assigned prior probabilities of occurrence, the
latter may be unequal. \cite{li} demonstrates, through numerical
simulation, that random texts follow Zipf's law because of the choice
of rank as the independent variable in that relationship. Therefore,
this paper concluded that Zipf's law is not an intrinsic property of
natural language of any major consequence. Recent work~\cite{Cancho10} 
%(Ferrer i Cancho and Elvevaag 2009, Ferrer-i-Cancho, R. and Gavald¨¤, R. 2009)
has shown that this is not the case.

In \cite {masuccirodgers}, the authors state that a linguistic network
formed from a random permutation of words also exhibits a
power-law. They attribute this to the nature of the linguistic network
created which is essentially a co-occurrence network. A co-occurrence
network, for our purposes, is a directed graph of words which are
linked if they are adjacent to each other in a sentence. The authors
reason that in such a network, the degree of a word is proportional to
its frequency within the text from which the network is
constructed. Thus, they say, permuting a text has no impact on the
degree distribution of a co-occurrence network. This is not true
unless duplicate edges can be added to the graph (which is clearly not
the case). An illustration is the dummy sentence \textit{A B C D E B
C} (degrees are \textit{A=1, B=3, C=2, D=2, E=2}) and one of its
random permutations \textit{D B E C B A C} (degrees are \textit{A=2,
B=4, C=3, D=1, E=2}). The identities of the words that co-occur with a
word, and whether the word itself occurs at the end or the beginning
of a sentence clearly affect its degrees in the graph.

In this work we analyze in detail the topology of 
the language networks at different levels of linguistic constraints.
We find that networks based on randomized text exhibit small-world
and scale-free characteristics. We also show that many of the network statistics we study
are functions of the distribution of the underlying data from which the network is built.
This study tries to shed light on several aspects of human language. Language is a complex structure 
where words act as simple elements that come together to form language. We try to understand why
is it rather easy to link such simple elements to form sentences, novels,books,..etc.
We are also interested in studying the structural properties of word networks
that would provide humans with easy and fast word production. 
We study such word networks at different levels of linguistics constraints 
to better understand the role of such constraints in terms of making mental
navigation through words easy. In this study, we conduct experiments with four different languages
(English, French, Spanish, and Chinese). This allows us to find out
the features of word networks that are common to all languages.
The experiments we will describe throughout this paper were able to find some answers
that will ultimately lead to the answers to those harder questions.

\begin{figure*}
\includegraphics[scale=0.60]{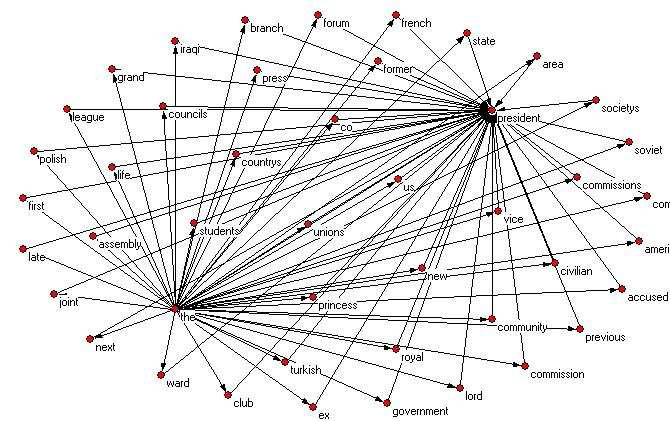}
\caption{\label{fig:one}Two hops from 'the' to 'president' in the frequent bigrams network}
\end{figure*}

We study the properties of linguistic networks at different levels of 
linguistic constraints. We study the networks formed by
randomizing the underlying text, and how they compare to the
properties of networks formed from plain English
Broadly, two types of networks are built in this paper - 
frequent bigrams (should we call these collocations?) and
co-occurrence networks. A frequent bigram can be defined as a sequence
of two or more words that is statistically idiosyncratic. That is, in
a certain context, the constituent words of a frequent bigram co-occur
with a significant frequency. In \cite{manningschutze}, a collocation
(frequent bigram) is defined as a conventional way of saying
something. \textit{Library card}, \textit{phone booth} and
\textit{machine translation} are examples of collocations. A frequent
bigrams network, simply, is a graph of words as vertices such that
there is an edge between two words provided they form a frequent
bigram. Frequent bigrams networks are built from unscrambled text for
the purpose of general study. Here, the method presented in
\cite{canchosole} is improved upon by employing a Fisher's exact test
to extract frequent bigrams \cite{pedersen}. A sample from the
frequent bigrams graph constructed in this study is shown in figure
\ref{fig:one}.

The contributions of this paper include: (1) we used uniformly
distributed permutation algorithms for scrambling the corpus and added
the Fisher exact test for detecting significantly correlated word
pairs, (2) we obtained good results on networks induced from random
corpus. Randomization of the corpus reduces linguistic
constraints. Linguistic constraints reduce the small-worldness of the
network!, (3) we used network randomization algorithms that
preserve the degree distribution, and (4) we used lemmatized words and
considered multiple languages.

The paper is organized as follows: in section \ref{sec:related}, we review some related work
and put our work in context with respect to related work. Section \ref{sec:method}
describes our methodology and how the different networks are created. Finally in section \ref{sec:results},
we analyze the different properties of all networks.

\section{Related Work}
\label{sec:related}
Network models of language have been studied by many researchers in
the past~\cite{canchosole,Steyvers05}. 

One of the earliest such studies \cite{canchosole} constructed
a ``restricted'' (frequent bigrams) network and an ``unrestricted''
(co-occurrence) network from a subset of the British National
Corpus~\cite{BNC}. These networks were
undirected. In that work, a bigram, $w_{i}w_{i+1}$, is treated as a
frequent bigram if the probability of its occurrence, $p_{i,i+1}$, is
found to be greater than
$p_{i}\cdot p_{i+1}$, under the assumption of the constituent words
being independent. This is a simplistic and unreliable filter for word
dependence \cite{manningschutze}. \cite{pedersen} demonstrates the
suitability of Fisher's (right-sided) Exact test for this purpose. In
the unrestricted network, words are linked if they co-occur in at
least one sentence within a window of two words. Both these networks
were shown to possess small-world characteristics with the average
minimum distance between vertices $\approx 3$. They were also shown to
be scale-free, with degree distributions following composite
power-laws with exponents $\gamma _1=-1.50$ and $\gamma _2=-2.70$
respectively. The latter exponent is approximately equal to $\gamma = -3$,
obtained for a random graph constructed by employing the
Barabasi-Albert (BA) model with preferential attachment. Also studied
was a ``kernel'' network of the 5000 most connected words extracted
from the restricted network. This network too was shown to follow a
power-law degree distribution with $\gamma =-3$. The authors go on to
suggest that power-law coefficients obtained are indicative of
language having evolved following the law of preferential
attachment. They also suggest that the kernel network constructed here
is representative of human mental lexicons; these lexicons possess the
small-world feature which facilitates quick navigation from one word
to another.

Beyond universal regularities such as Zipf's law, \cite{SFM} recently
examined burstiness, topicality, semantic similarity distribution and
their interrelation and modeled them with two mechanisms, namely
frequency ranking with dynamic reordering and memory across
documents. Besides, large web datasets were used to validate the
model. This paper and several other papers focused on modeling human
written text with specific mechanisms.

Network theory has been used in several studies about
the structure of syntactic dependency networks. In \cite{ferrerpql},
the author overview-ed the past studies on linguistic networks and discussed
the possibilities and advantages of network analysis of syntactic
dependency networks.  In \cite{fsk}, network properties such as small
world structure, heterogeneity, hierarchical organization, betweenness
centrality and assortativeness etc were examined for the syntactic
networks from Czech, Romanian and German corpora.  Seven corpora were
examined by similar complex network analysis methods in
\cite{fmpd}. Several common patterns of syntactic dependency networks
were found. These patterns include high clustering coefficient of each
vertex, the presence of a hierarchical network organization,
disassortative mixing of vertices.  In \cite{fcc}, spectral methods
were introduced to cluster the words of the same class in a syntactic
dependency network.

In \cite{shuffle}, the authors examined the structural properties of
two weighted networks, a linguistic network and a scientific
collaboration network. The linguistic network in this paper is simply
a co-occurrence network instead of syntactic dependency network. The
weight of edges between vertices were considered in the paper. The
networks built from shuffled text were used as the null hypothesis to
compare with the real network in order to find the characteristic of
the real ones. Through the analysis of differences between the real
network and a shuffled network, they proved that the scale free degree
distribution are induced by Zipf's law.

In \cite{dorogovstevmendes}, the authors model the evolution of a network of language based upon preferential attachment. That is, a new word is connected to a word in the network with probability proportional to the latter's degree. The model that they develop, almost astonishingly, agrees very well with the empirical results obtained in \cite{canchosole}. The degree distribution of the theoretical network follows a composite power-law with exponents exactly equal to those obtained by \cite{canchosole}. This work further validates the scale-free nature of human language.

A stochastic model of language was created on the basis of combined \textit{local} and \textit{global} preferential attachment (PA) in \cite{masuccirodgers}. A new word attaches to the nearest neighbor with highest degree in the local PA scheme, whereas it attaches to the word with the highest degree in the entire graph in global PA. They find that their model produces a network with scale-free properties with various statistics in agreement with empirical evidence. They argue that plain PA schemes don't necessarily take into account the hierarchical nature of networks as is hinted at by Zipf's law and hence their combined model, which follows a mixed local-global growth is more suitable for the desired purpose.

\begin{figure}
\includegraphics[scale=0.60]{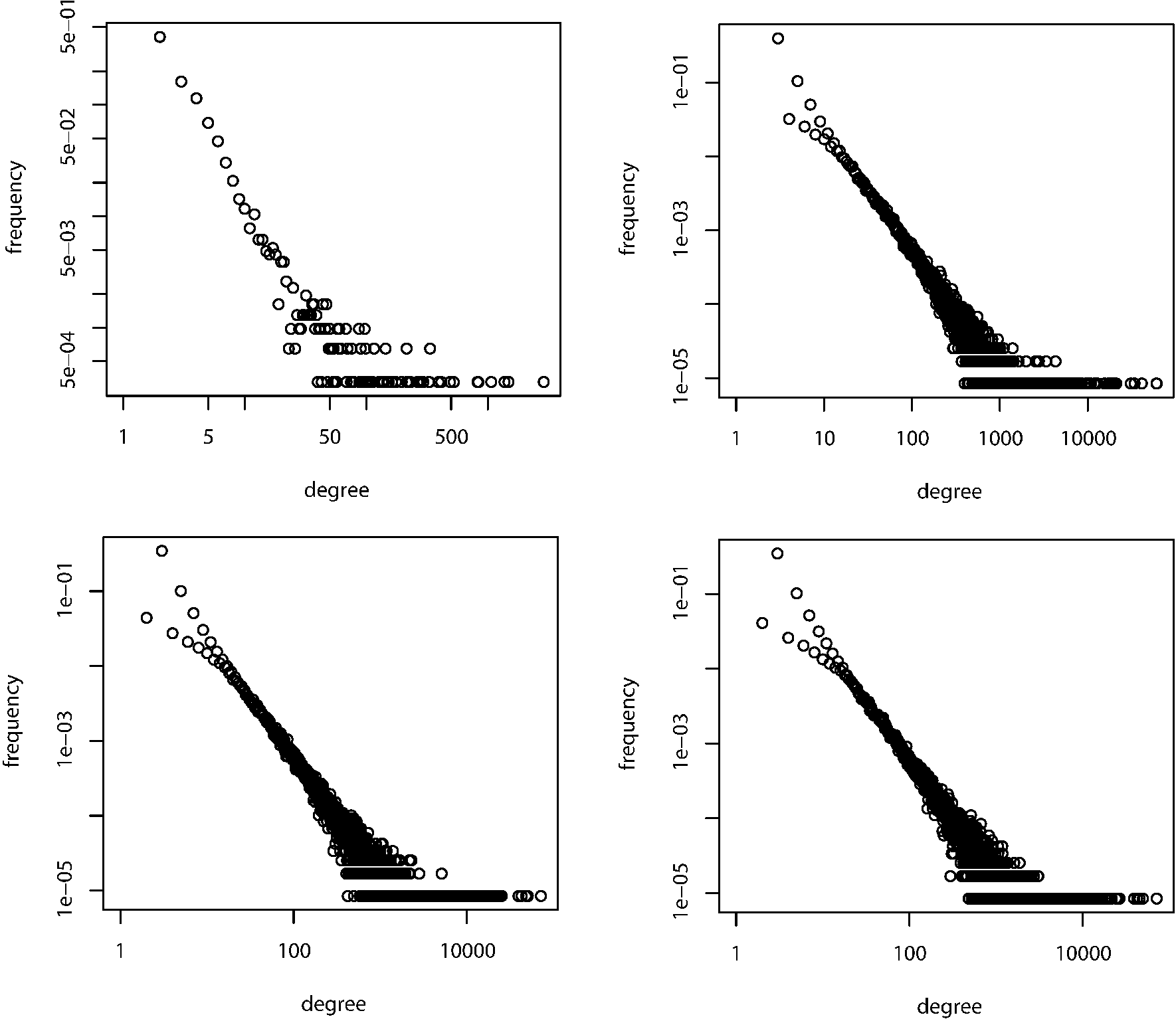}
\caption{\label{fig:two}Degree distributions for ENG-COLL, ENG, ENG-RANSEN and ENG-RANDOC}
\end{figure}

\begin{figure}
\centering
\includegraphics[scale=0.40]{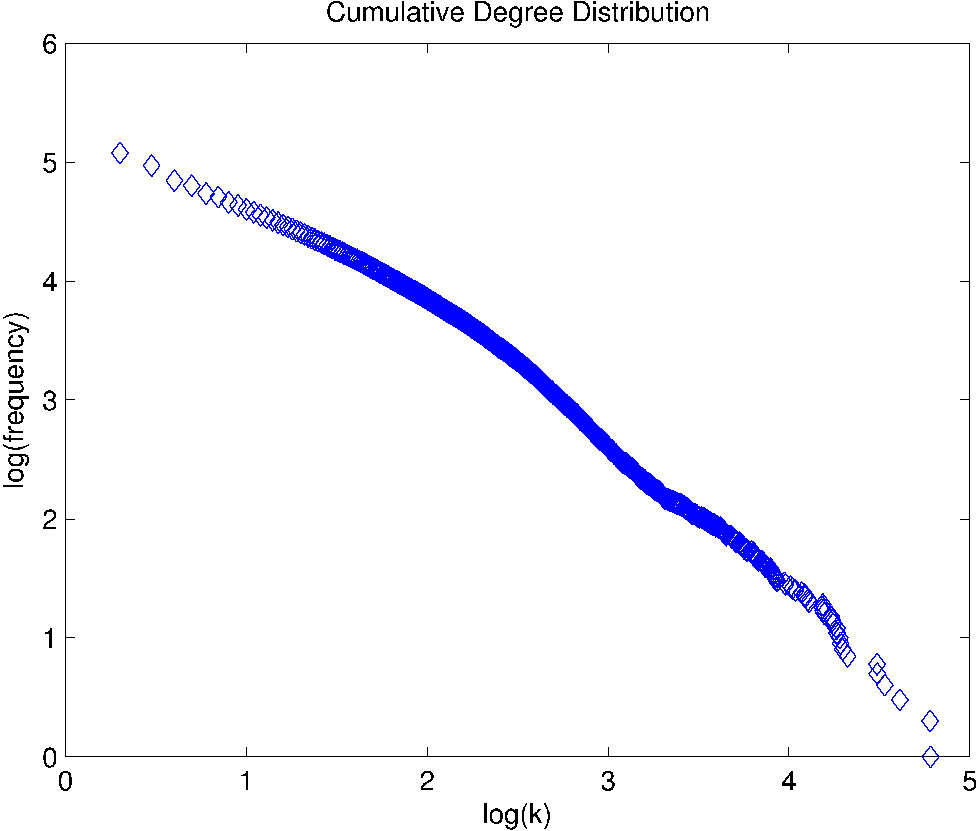}
\caption{\label{fig:three}Cumulative degree distribution for ENG}
\end{figure}

\section{Methodology}
\label{sec:method}
In this article we consider the two aspects that have not been previously studied in depth in word cooccurrence networks: (1) the effect of the strength of links and (2) the word forms that vertices stand for.

We consider four levels of cooccurrence constraints (from less to more constraints):
\begin{enumerate}
\item
RANDOC \\ 
Words were permuted randomly within each document. Two vertices are linked if the corresponding words are adjacent in the permuted text. 
sentence boundaries were respected. That is, the length of the original sentences was kept constant 
\item
RANSEN \\
Words were permuted randomly within each sentence. Two vertices are linked if the corresponding words are adjacent in the permuted text.
\item
PLAIN \\
Two vertices are linked if the corresponding words are adjacent in the original text.
\item
COLL \\
Two vertices are linked if the corresponding words form a highly associative bigram.
Fisher's Exact Test was used to extract highly associative bigrams (significance value $\leq 0.01$) from the unscrambled English text. 
Fisher's Exact Test is used because it is considered a more suitable test for determining word associativeness \cite{pedersen}
We assume that the frequency associated with a bigram $<word1><word2>$ is stored in a 2x2 contingency table:

\begin{center}
\begin{tabular*}{0.5\textwidth}{@{\extracolsep{\fill}}ccc}
\\
 & word2 & $\neg$word2\\
word1 & n11 & n12\\
$\neg$word1 & n21 & n22\\
\\
\end{tabular*}
\end{center}

where n11 is the number of times $<word1><word2>$ occur together, 
n12 is the number of times $<word1>$ occurs with some word other than
$word2$, and so on. Fisher's exact test is calculated by fixing the marginal
totals and computing the hypergeometric probabilities for all the
possible contingency tables.

\end{enumerate}

We also consider two kinds of vertices
\begin{enumerate}
\item
RAW \\
Words are used in their raw forms. 
\item
LEMM \\
A lemmatized form of words is used.
\end{enumerate}

For each corpus, a different cooccurrence network is built for each level of constraints and type of vertex. 
This results in 8 different networks for each language.
We assume that that our graphs are undirected and that loops are allowed 

We define the structure of an undirected graph of $n$ vertices through a binary adjacency matrix $A = \left\{ a_{ij} \right\}$, where $a_{ij}=1$ if the vertex $i$ and the vertex $j$ are linked and $a_{ij}=0$ otherwise. Notice that the matrix is symmetric ($a_{ij}=a_{ik}$) and $a_{ii}=1$ is possible.
We define $k_i$, the degree of the $i$-th vertex, as  
\begin{equation}
   k_i = \sum_{j=1}^n a_{ij}.
\end{equation}

The English text we used to construct all the networks comes from a random subset of the British National Corpus.
The subset had 7.5M words. We used the Fisher-Yates shuffle algorithm for creating random permutations within sentences and documents.
This algorithm produces each possible permutation with equal probability.

We report several general network statistics including average degree, diameter,
average shortest path, global and local clustering coefficient. The diameters and average shortest paths
were calculated using the maximum connected components while the other statistics were calculated used the whole network.

In addition to English, we consider several other languages.
We used a subset of the Spanish, French, and Chinese Gigaword corpora.
We built four different networks for each language (RANDOC, RANSENT, PLAIN, and COLL).
More statistics regarding the datasets used for constructing the networks is shown in Table~\ref{tab:corpus}.

\begin{figure}
\centering
\includegraphics[scale=0.30]{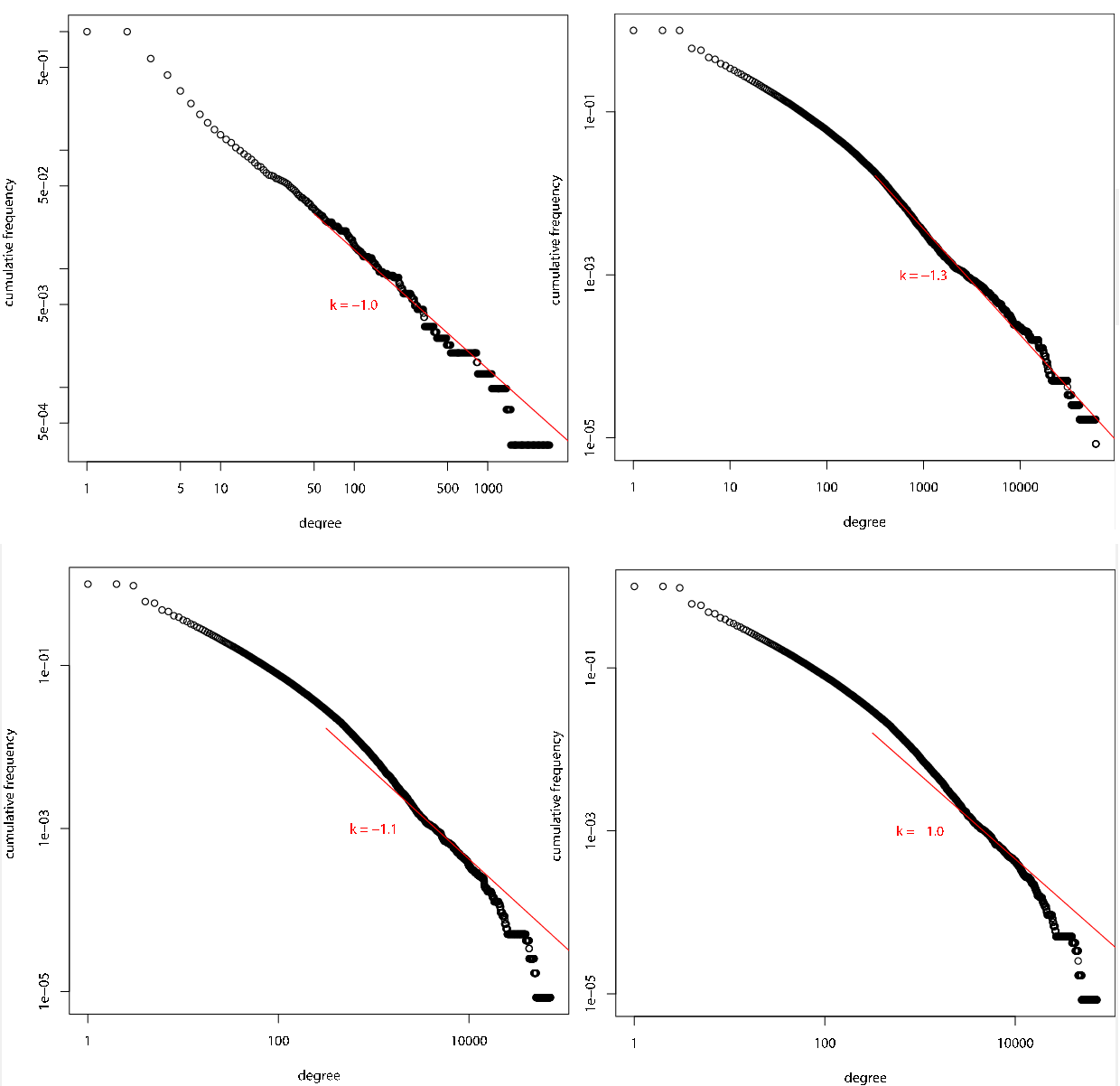}
\caption{\label{fig:four}Line fitted to the tail of the cumulative degree distribution for ENG-COLL, ENG, ENG-RANSEN, ENG-RANDOC}
\end{figure}

\begin{table}[ht]
\caption{\label{tab:fre} Top ten frequent words in English corpora}
\centering
\begin{tabular*}{0.25\textwidth}{@{\extracolsep{\fill}}cc}
Word & Frequency\\
\hline
the & 503322\\
of & 246620\\
to & 213108\\
and & 208294\\
a & 182067\\
in & 157755\\
is & 79274\\
that & 79265\\
was & 73143\\
for & 71614\\
\end{tabular*}
\end{table}

\begin{table}[ht]
\caption{\label{tab:colltopten} Top ten most connected words in ENG-COLL}
\centering
\begin{tabular*}{0.25\textwidth}{@{\extracolsep{\fill}}cc}
Word & Degree\\
\hline
the & 2505\\
of & 1492\\
to & 1408\\
a & 1095\\
in & 913\\
and & 795\\
is & 522\\
was & 483\\
that & 458\\
it & 436\\
\end{tabular*}
\end{table}

\begin{table}[ht]
\caption{\label{tab:engtopten} Top ten most connected words in ENG}
\centering
\begin{tabular*}{0.25\textwidth}{@{\extracolsep{\fill}}cc}
Word & Degree\\
\hline
the & 60808\\
and & 60514\\
of & 41141\\
in & 33967\\
to & 30897\\
a & 30810\\
for & 21178\\
with & 19912\\
is & 19303\\
as & 19158\\
\end{tabular*}
\end{table}

\begin{table}[ht]
\caption{\label{tab:ransentopten} Top ten most connected words in ENG-RANSEN}
\centering
\begin{tabular*}{0.25\textwidth}{@{\extracolsep{\fill}}cc}
Word & Degree\\
\hline
the & 71498\\
of & 50635\\
and & 48239\\
to & 42940\\
a & 42750\\
in & 39543\\
is & 25264\\
for & 24556\\
that & 23786\\
was & 23703\\
\end{tabular*}
\end{table}

\begin{table}[ht]
\caption{\label{tab:randoctopten} Top ten most connected words in ENG-RANDOC}
\centering
\begin{tabular*}{0.25\textwidth}{@{\extracolsep{\fill}}cc}
Word & Degree\\
\hline
the & 71708\\
of & 49673\\
to & 45677\\
and & 45665\\
a & 42779\\
in & 39332\\
is & 26648\\
that & 26143\\
for & 25133\\
was & 24696\\
\end{tabular*}
\end{table}

\begin{table}
\centering
\begin{tabular}{cccc}
Language & Corpus length (in kilo words) & Kind of vertex & Number of different vertices \\
ENG & 7500  & RAW & $118889$ \\
ENG & 7500  & STEMMED & $71521$ \\
FRE & 1700  & RAW & $49932$ \\
FRE & 1700  & STEMMED & $33589$ \\
SPA & 1300  & RAW & $61260$ \\
SPA & 1300  & STEMMED & $29216$ \\
CHI & 170 & - & $4573$
\end{tabular}
\caption{\label{tab:corpus} Corpus length is the length of the dataset used for constructing the network which may not coincide with the length of the whole corpus (the parenthesis show the proportion of the real corpus used).}
\end{table}

\section{Results}
\label{sec:results}

\begin{figure}
\centering
\includegraphics[scale=0.20]{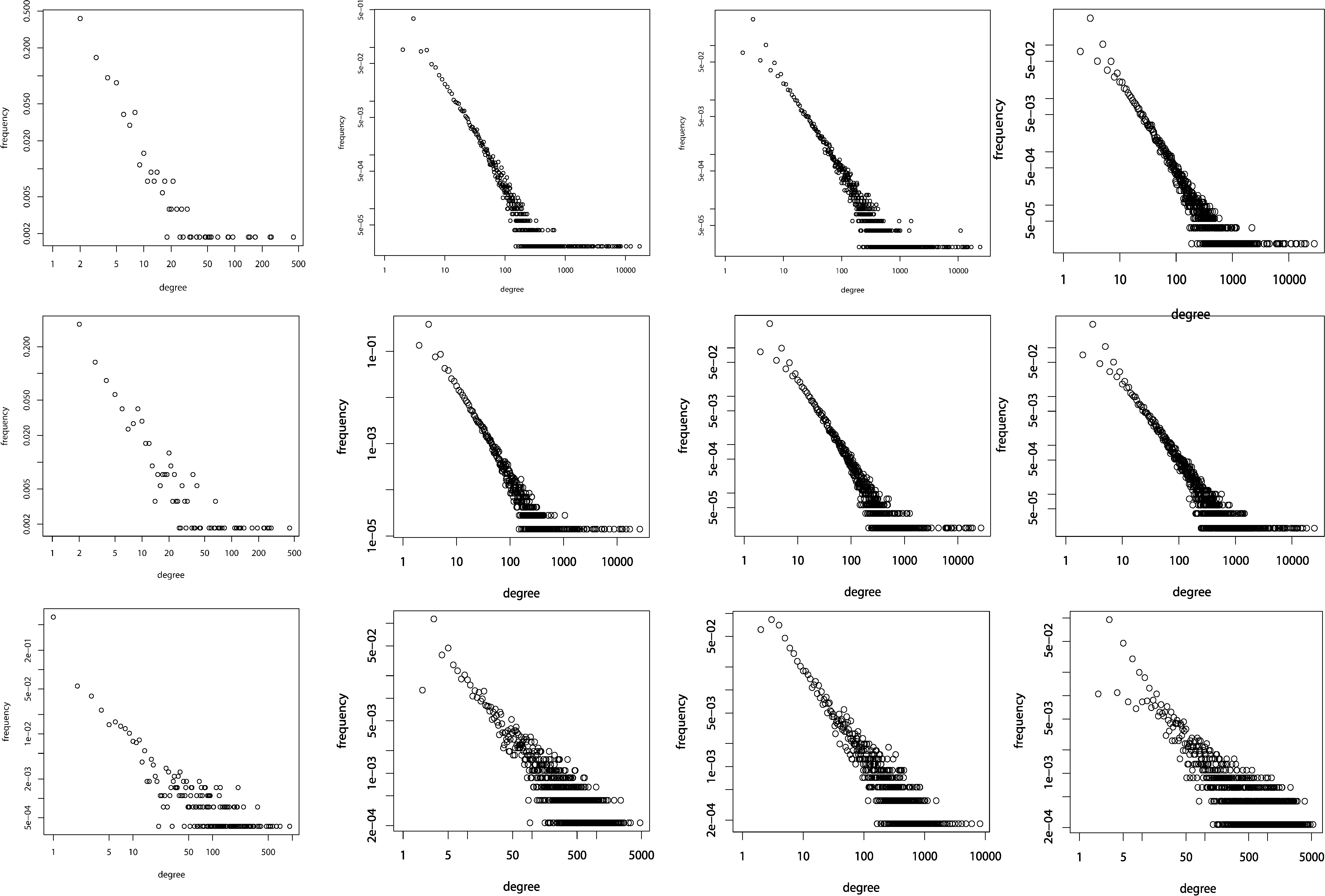}
\caption{\label{fig:six}Degree distribution for SPA-COLL, SPA, SPA-RANSEN, SPA-RANDOC, FRE-COLL, FRE, FRE-RANSEN, FRE-RANDOC, CHI-COLL, CHI, CHI-RANSEN, CHI-RANDOC}
\end{figure}

\begin{figure}
\centering
\includegraphics[scale=0.2]{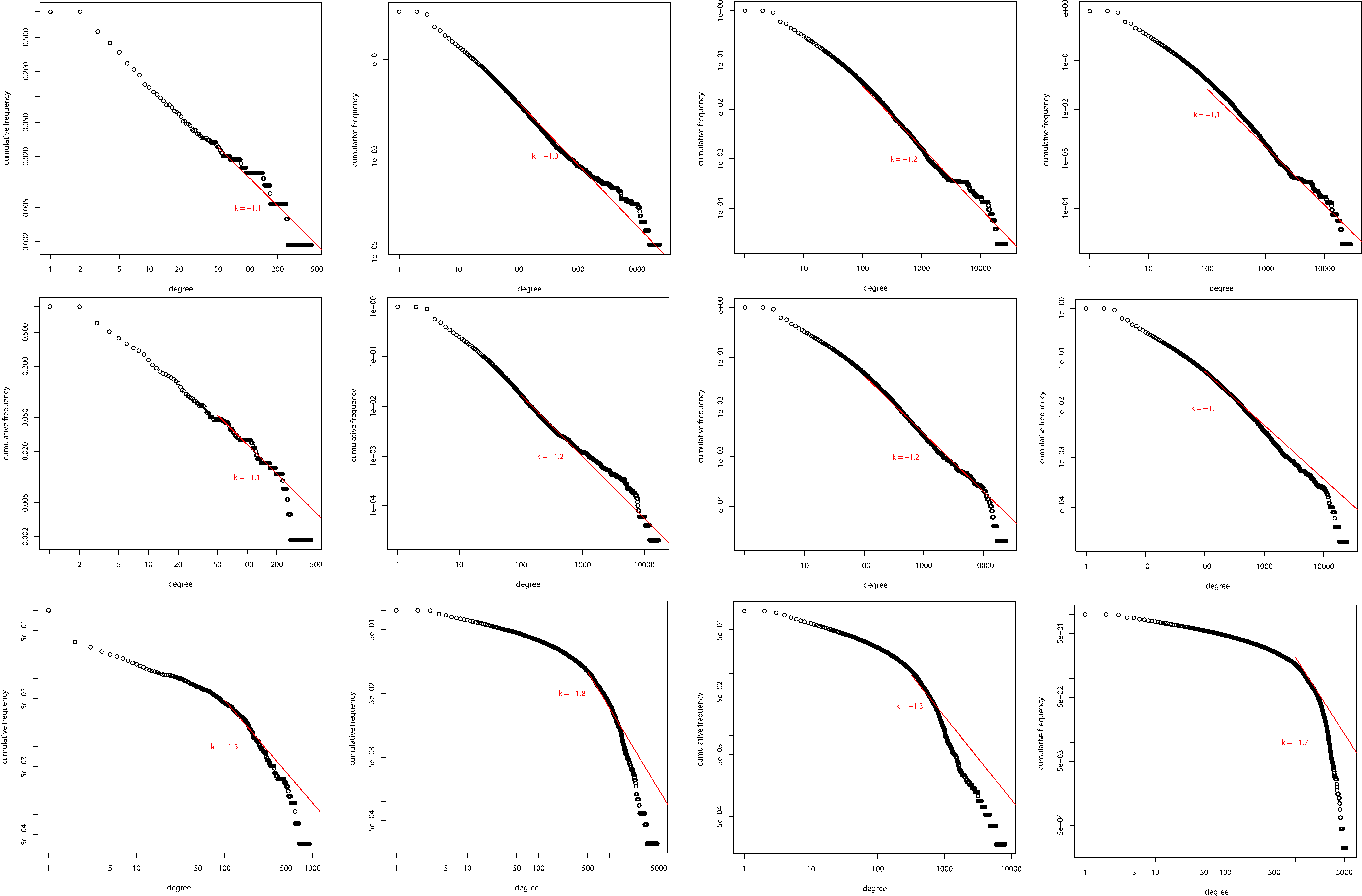}
\caption{\label{fig:seven}Cumulative degree distribution for SPA-COLL, SPA, SPA-RANSEN, SPA-RANDOC, FRE-COLL, FRE, FRE-RANSEN, FRE-RANDOC, CHI-COLL, CHI, CHI-RANSEN, CHI-RANDOC}
\end{figure}

\begin{figure}
\centering
\includegraphics[scale=0.50]{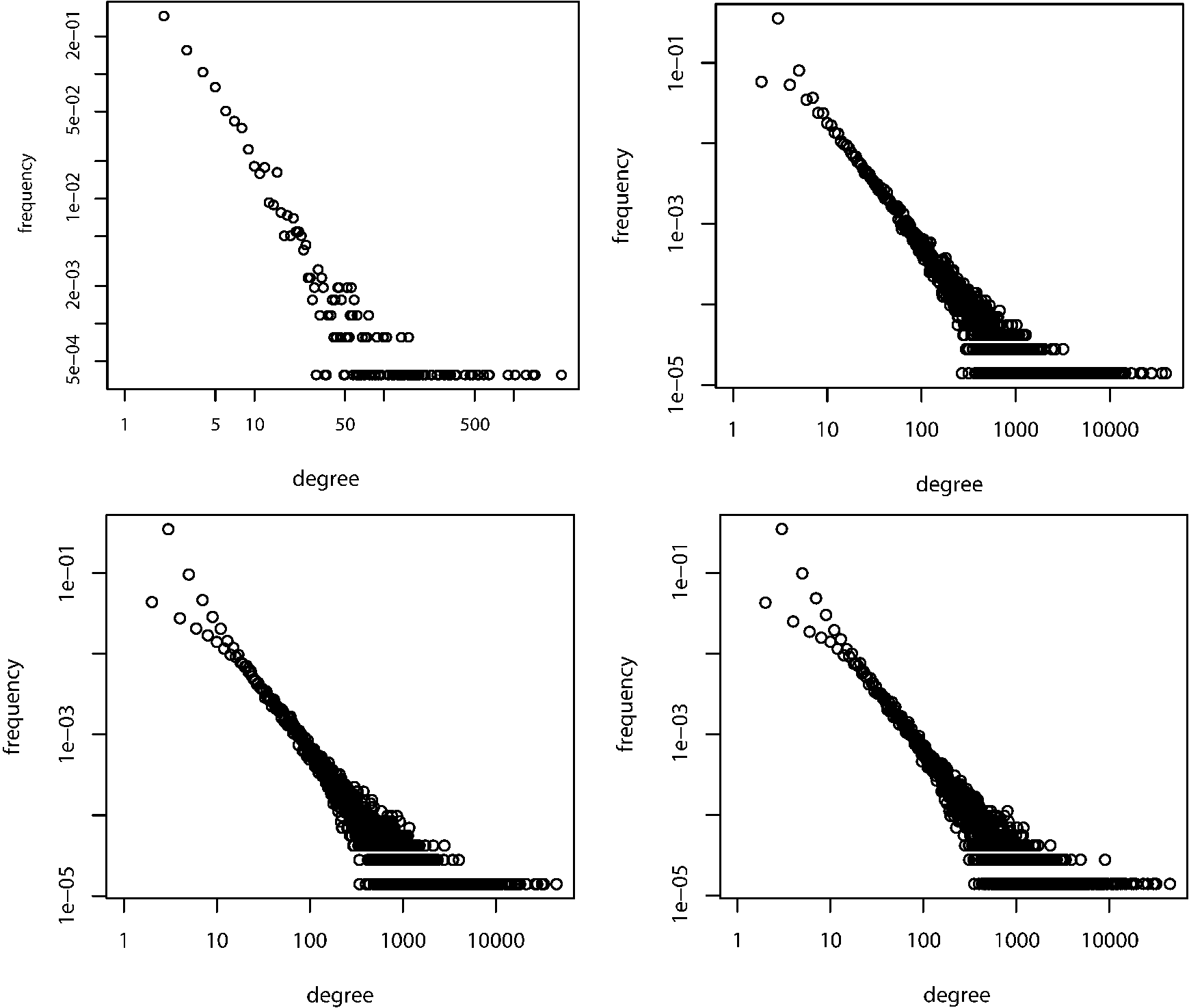}
\caption{\label{fig:nine}Degree distribution for stemmed ENG-COLL, ENG, ENG-RANSEN and ENG-RANDOC}
\end{figure}

\begin{figure}
\centering
\includegraphics[scale=0.30]{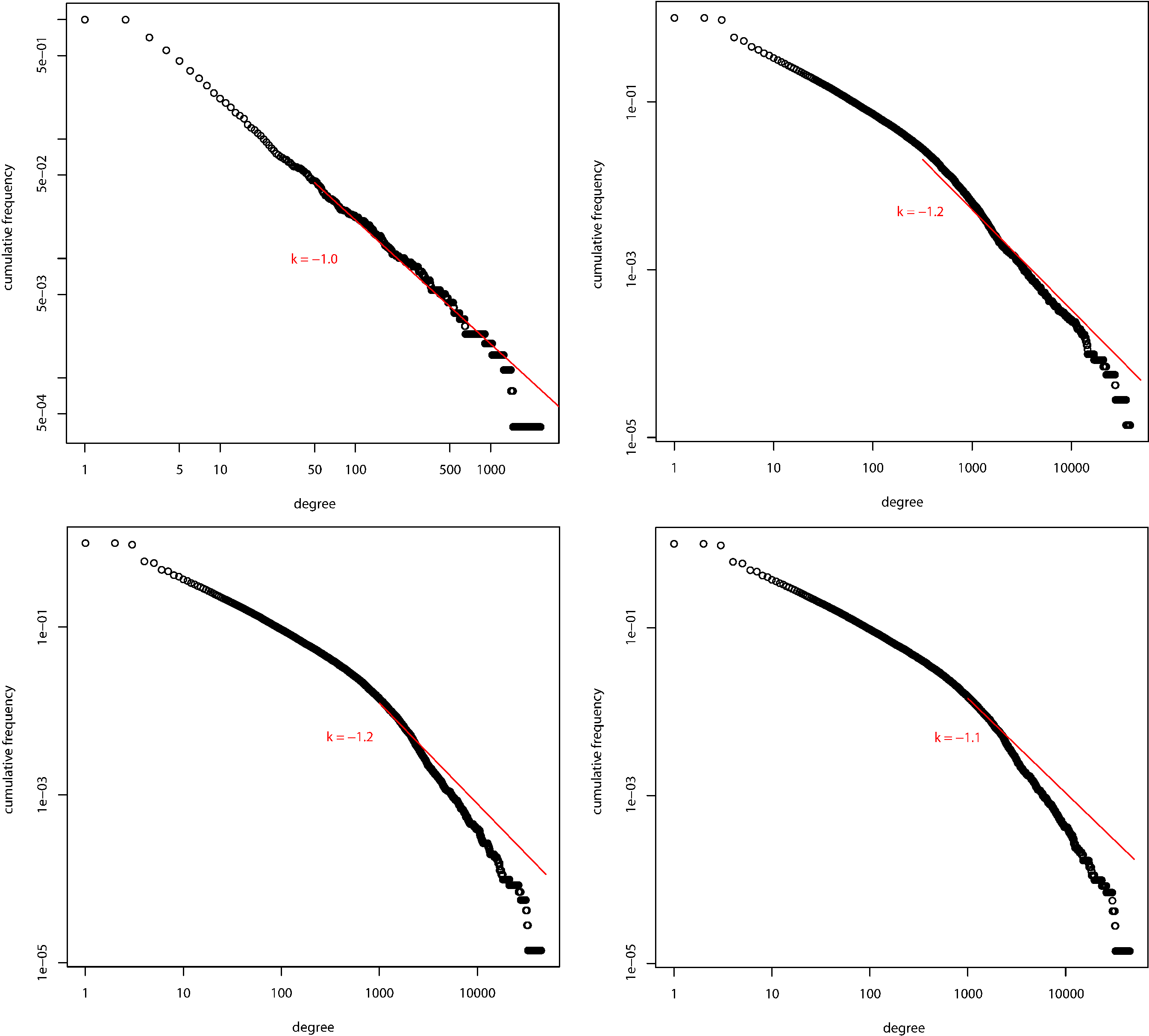}
\caption{\label{fig:ten}Cumulative degree distribution for stemmed ENG-COLL, ENG, ENG-RANSEN and ENG-RANDOC}
\end{figure}

\begin{figure}
\centering
\includegraphics[scale=0.50]{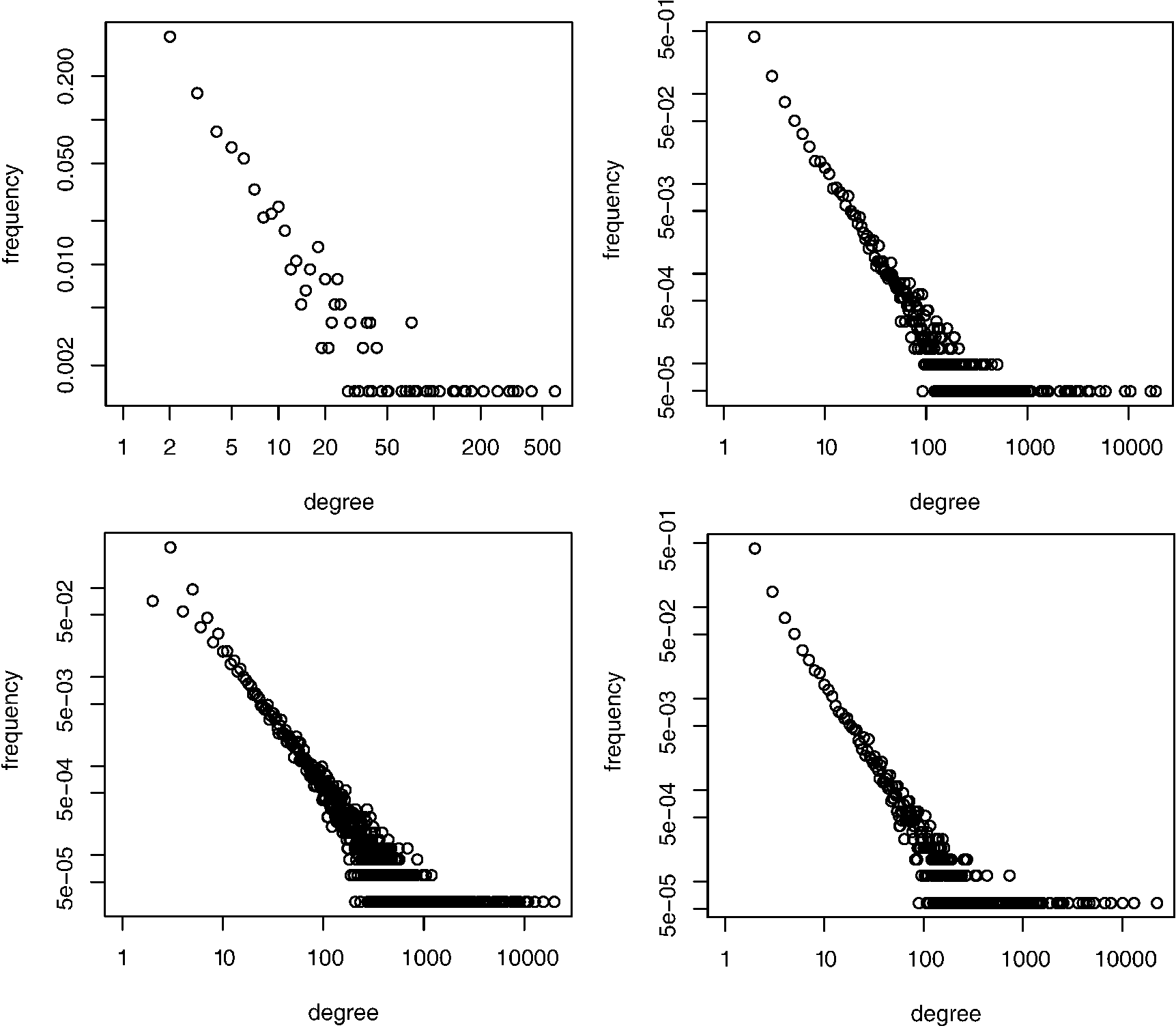}
\caption{\label{fig:eleven}Degree distribution for stemmed FRE-COLL, FRE, FRE-RANSEN and FRE-RANDOC}
\end{figure}

\begin{figure}
\centering
\includegraphics[scale=0.50]{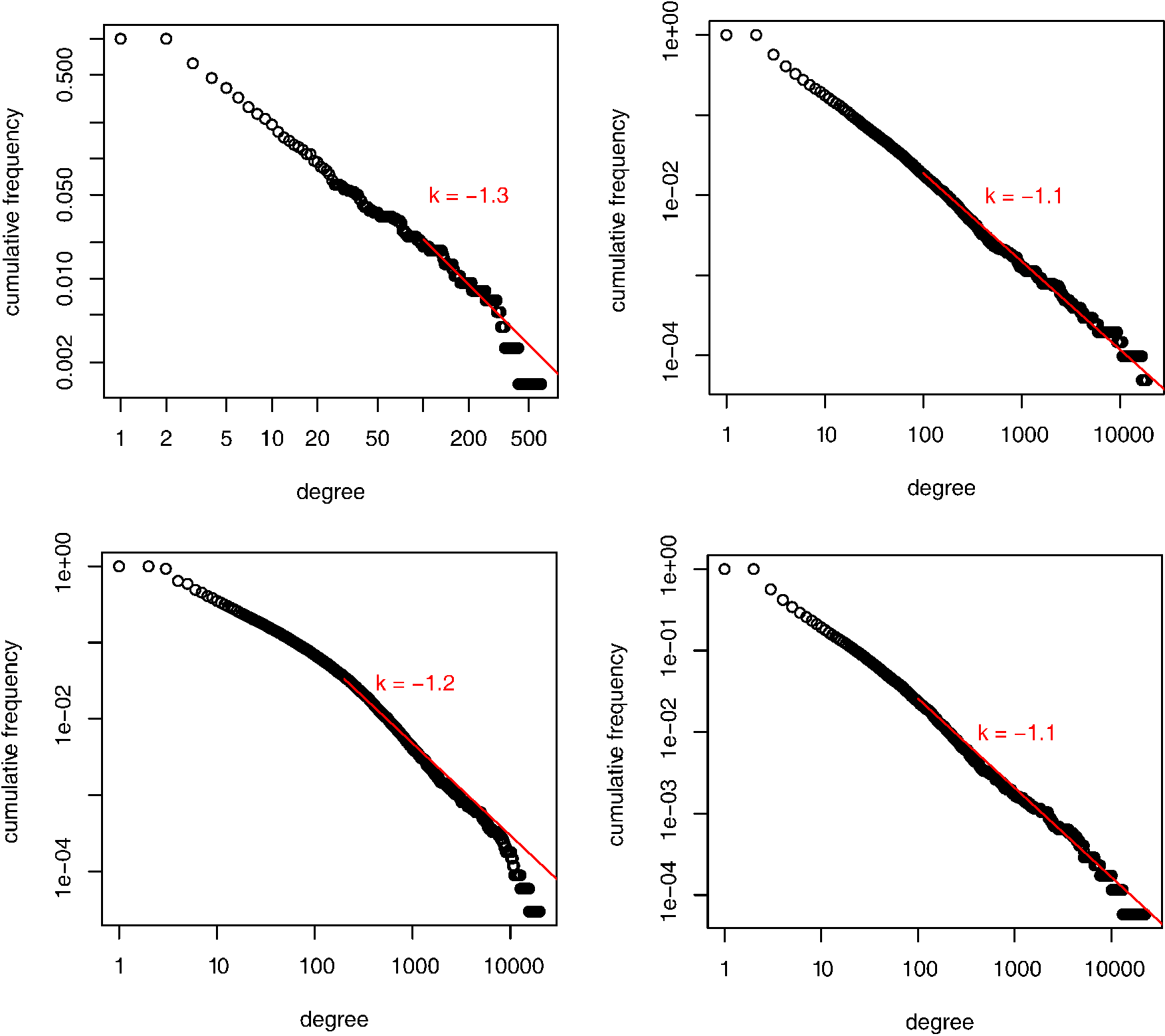}
\caption{\label{fig:twelve}Cumulative degree distribution for stemmed FRE-COLL FRE, FRE-RANSEN and FRE-RANDOC}
\end{figure}

\begin{figure}
\centering
\includegraphics[scale=0.30]{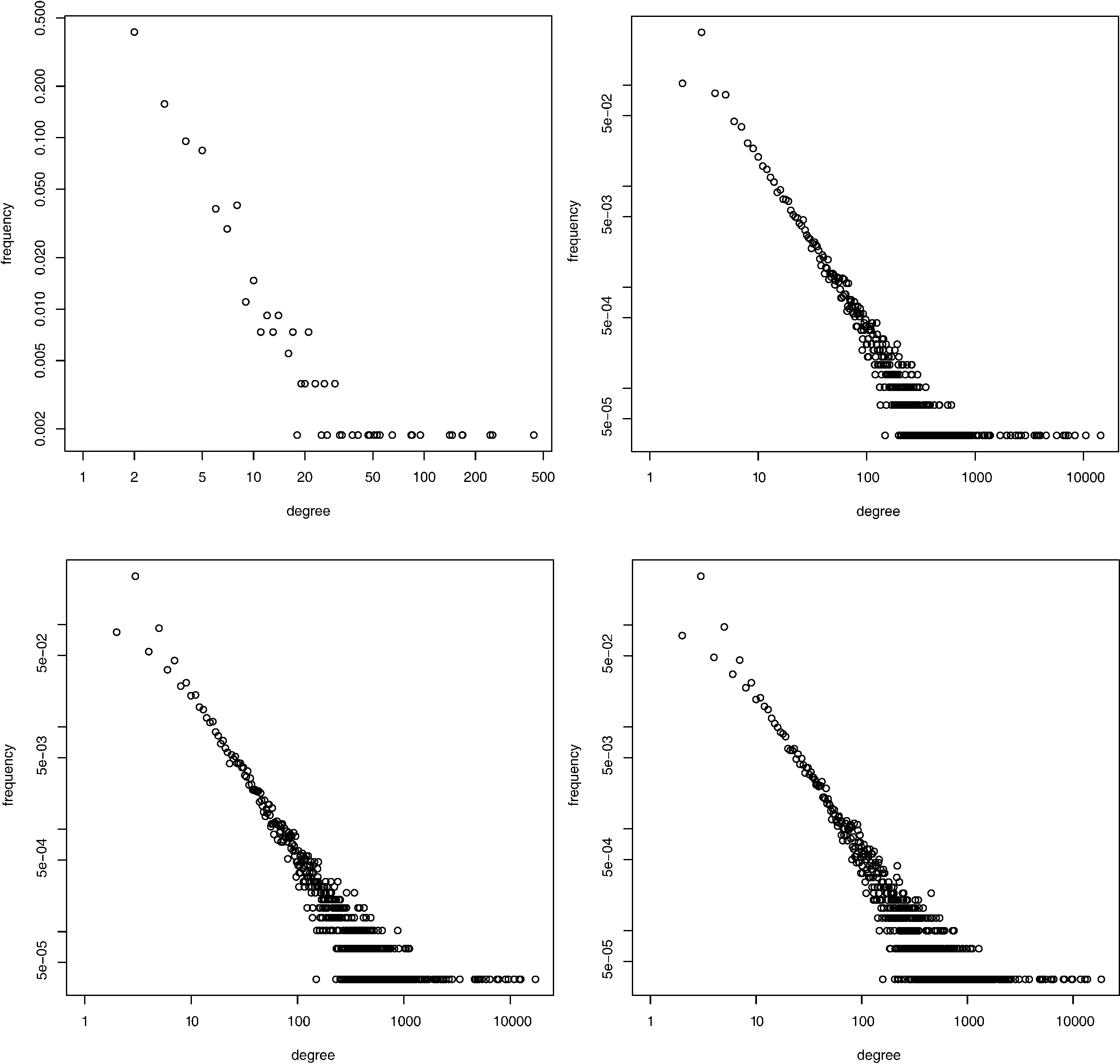}
\caption{\label{fig:thirteen}Degree distribution for stemmed SPA-COLL, SPA, SPA-RANSEN and SPA-RANDOC}
\end{figure}

\begin{figure}
\centering
\includegraphics[scale=0.30]{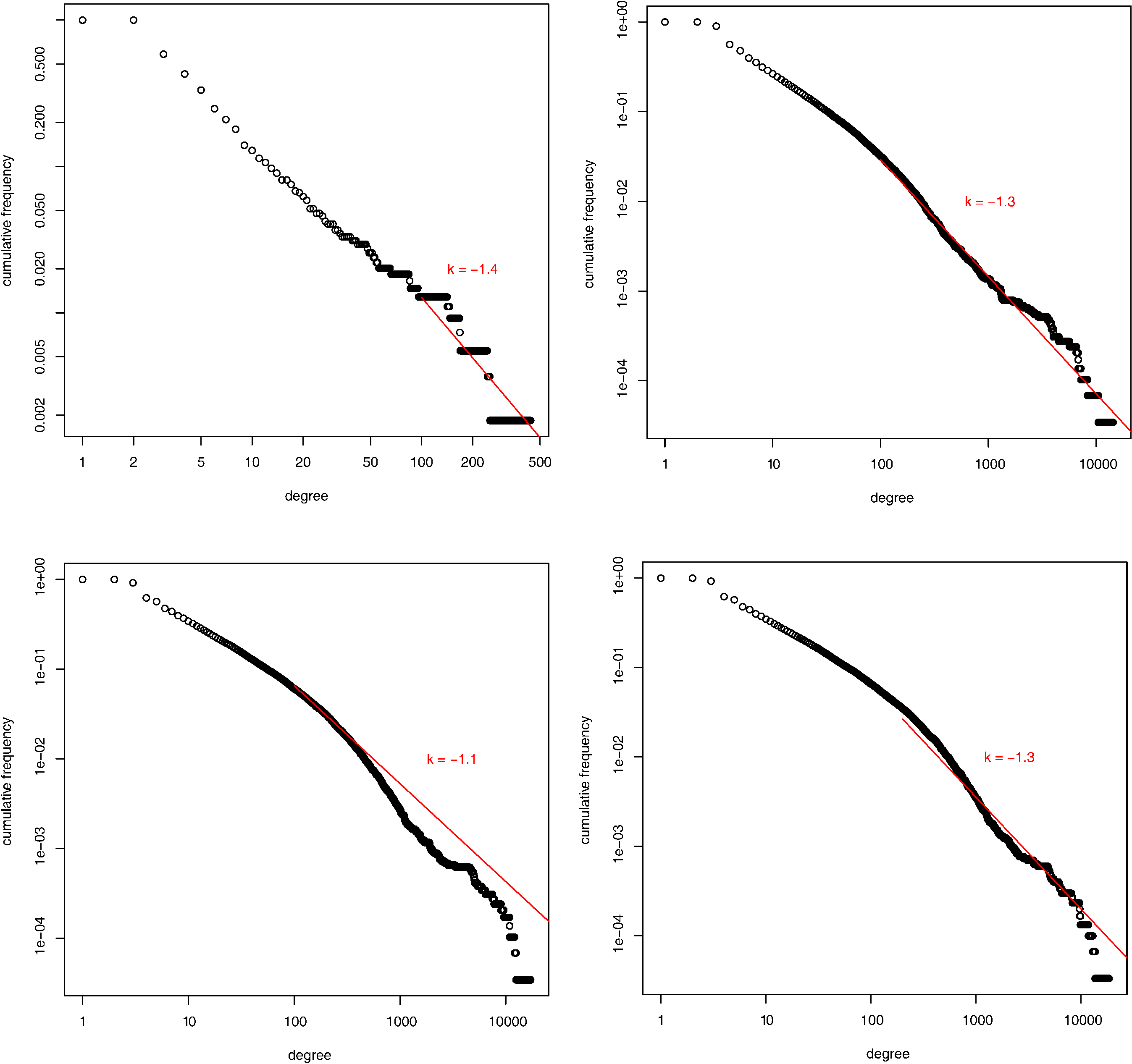}
\caption{\label{fig:fourteen}Cumulative degree distribution for stemmed SPA-COLL, SPA, SPA-RANSEN and SPA-RANDOC}
\end{figure}

\subsection{Degree Distribution}
The degree distributions for all four networks ENG-COLL, ENG, ENG-RANSEN and ENG-RANDOC
are plotted in figure \ref{fig:two}. It is worth mentioning that
all figures are on a base 10 log-log scale. All four
distributions show characteristics of a power-law. The tails of the
corresponding cumulative degree distributions are plotted in figure
\ref{fig:four} and are fitted by lines. 
%The thresholds over which to
%fit were decided by simply observing the data ($50$ for ENG-COLL and
%$10^{2.5}$ for ENG, ENG-RANSEN, ENG-RANDOC). 
The power-law coefficient is
fitted with the maximum likelihood method as recommended in \cite{newman}. 
The coefficients thus obtained from the plots are
$\gamma_{ENG-COLL}=2.0$, $\gamma_{ENG}=2.3$, $\gamma_{ENG-RANSEN}=2.1$ and
$\gamma_{ENG-RANDOC}=2.0$. These are all very similar values and are also
close to the values obtained for co-occurrence and frequent bigrams
($\gamma =2.7$ each) networks in \cite{canchosole}.

For Spanish, French and Chinese experiment, the degree distributions
are plotted in Figure. \ref{fig:six}. The corresponding cumulative
degree distributions are plotted in Figure. \ref{fig:seven}. All distributions show
characteristics of a power-law. The coefficients for the networks are
$\gamma_{SPA-COLL}=2.1$, $\gamma_{FRE-COLL}=2.1$, $\gamma_{CHI-COLL}=2.5$
$\gamma_{SPA}=2.3$, $\gamma_{FRE}=2.2$, $\gamma_{CHI}=2.8$,
$\gamma_{SPA-RANSEN}=2.2$, $\gamma_{FRE-RANSEN}=2.2$, $\gamma_{CHI-RANSEN}=2.3$,
$\gamma_{SPA-RANDOC}=2.1$, $\gamma_{FRE-RANDOC}=2.1$, $\gamma_{CHI-RANDOC}=2.7$.
These are similar to the English dataset
($\gamma_{ENG}=2.3$, $\gamma_{RANSEN}=2.1$).

The distributions of the networks based on the lemmatized form of the words are
pretty similar to the original ones. The degree distributions for
stemmed ENG-COLL, ENG and ENG-RANSEN are plotted in Figure~\ref{fig:nine} and Figure~\ref{fig:ten}.
The coefficients are $\gamma_{STEMENG-COLL}=2.0$, $\gamma_{STEMENG}=2.2$, $\gamma_{STEMENG-RANSEN}=2.2$ and $\gamma_{STEMENG-RANDOC}=2.1$.
The degree distributions for
stemmed FRE-COLL, FRE, FRE-RANSEN and FRE-RANDOC are plotted in Figure~\ref{fig:eleven} and Figure~\ref{fig:twelve}.
The coefficients are $\gamma_{STEMFRE-COLL}=2.3$, $\gamma_{STEMFRE}=2.1$, $\gamma_{STEMFRE-RANSEN}=2.2$ and $\gamma_{STEMFRE-RANDOC}=2.1$.
The degree distributions for
stemmed SPA-COLL, SPA, SPA-RANSEN and SPA-RANDOC are plotted in Figure~\ref{fig:thirteen} and Figure~\ref{fig:fourteen}.
The coefficients are $\gamma_{STEMSPA-COLL}=2.4$, $\gamma_{STEMSPA}=2.3$, $\gamma_{STEMSPA-RANSEN}=2.1$ and $\gamma_{STEMSPA-RANDOC}=2.3$.
%The network statistics are shown in table \ref{tab:six}, \ref{tab:seven} and \ref{tab:eight}.

\subsection{Link density}

\begin{table*}[ht]
\caption{\label{tab:two} Mean Degree ($ \bar{k}$) for all Networks}
\centering
\begin{tabular}{ccccc}
Statistic & COLL & PLAIN & RANSEN & RANDOC\\
\hline
ENG-RAW  & 0.32 & 37.18 & 52.97 & 55.50\\
ENG-LEMM & 0.53 & 45.54 & 70.88 & 74.85\\
FRE-RAW  & 0.13 & 15.70 & 29.65 & 33.36\\
FRE-LEMM & 0.24 & 20.24 & 38.70 & 44.07\\
SPA-RAW  & 0.06 & 12.27 & 22.96 & 25.42\\
SPA-LEMM & 0.17 & 19.67 & 33.24 & 36.59\\
CHI      & 8.41 & 171.07 & 114.43 & 436.01\\
\end{tabular}
\end{table*}

%\begin{enumerate}
%\item
As expected, the more cooccurrence constraints, the lower the density of links. In particular, we find a perfect negative rank correlation between the mean degree of vertices and the level of constraints for all the networks of the same language and the same kind of vertex. More formally, 
we have that the mean degree $\bar{k}$ obeys 
  $\bar{k}_{COLL} < \bar{k}_{PLAIN} < \bar{k}_{RANSEN} < \bar{k}_{RANDOC}$. 
Knowing that there are not ties between values of $\bar{k}$ for the same language and kind of vertex, the probability that the expected ordering is produced by chance is $1/4! \approx 0.041$. Thus, the perfect correlation between the level of constraints is statistically significant at a significance level of $0.05$.
  
Since the number of vertices of the network is the same for all networks of the same language (corpus) and the same kind of vertex, this perfect negative correlation is also equivalent to a perfect negative correlation between link density and level of constraints. The link density $\delta$ of a network where loops are allowed is defined as the proportion of linkable pairs of vertices that are linked. For the particular case of an undirected network where loops are allowed, we have 
\begin{eqnarray}
\delta & = & \frac{1}{{n+1 \choose 2}} \sum_{i=1}^n \sum_{j=i}^n a_{ij}, \\
       & = & \frac{\bar{k}}{n+1}, 
\end{eqnarray}
where $n$ is the number of vertices (if loops where not allowed this would be $\delta = \bar{k}/(n-1)$.
%\item
   The link density of the network of raw words is smaller than that of the network of stemmed words with the same level of constraints, i.e. $\bar{k} < \bar{k}_{STEMMED}$ or equivalently, $\delta < \delta_{STEMMED}$.
%\end{enumerate}

\subsection{Small-worldness}

We define $d$ as the shortest distance between a pair of nodes. 
The average shortest path is the average of all such distances.
The diameter of a network is the number of links in the shortest path between the furthest pair of nodes.
The diameter and the average shortest path for all networks
are shown in Table~\ref{tab:diameter} and Table~\ref{ref:shortpath} respectively.
The diameters for all four networks, as shown in Table~\ref{tab:diameter}, are small. These networks are small worlds. What is
surprising is that the diameters for RANSEN and RANDOC are smaller than that for ENG. This may be attributed
to the randomization produces which forms links between words which are otherwise not connected in ENG. This
results in a network that is even faster to navigate.

The mean shortest vertex-vertex distance, $\bar{d}$ obeys, 
  \begin{enumerate}
  \item
  $\bar{d} > \bar{d}_{STEMMED}$ for all languages and constrain levels.
  \item
  $\bar{d}_{COLL} < \bar{d}_{RANDOC} < \bar{d}_{RANSEN} < \bar{d}_{PLAIN}$
  in all languages except Chinese, where we have $\bar{d}_{RANDOC} < \bar{d}_{COLL} < \bar{d}_{RANSEN} < \bar{d}_{PLAIN}$. 
  \end{enumerate}

The diameter, i.e. the longest shortest path, $d^{max}$ obeys 
  \begin{enumerate}
  \item
  $d^{max} \geq d^{max}_{STEMMED}$ for all languages and constraint levels (we would have $d^{max} > d^{max}_{STEMMED}$ if COLL was excluded). 
  \item 
  $d^{max}_{COLL} < d^{max}_{RANDOC} < d^{max}_{RANSEN} < d^{max}_{PLAIN}$
  \end{enumerate}

%Dragomir: I am worried about the exception of Chinese for $\bar{d}_{COLL} < \bar{d}_{RANDOC} < \bar{d}_{RANSEN} < \bar{d}_{PLAIN}$ (see also general remarks). Please, put the formula that you use for computing $\hat{d}$ Do you include in the pair formed by a vertex and itself? What do you do with vertices with loops: the default assumption is that vertex is at minimum distance $0$ of itself, even if has a loop to itself. 

\begin{table*}[ht]
\caption{\label{tab:diameter} Diameters ($d^{max}$) for all networks}
\centering
\begin{tabular}{ccccc}
Statistic & COLL & PLAIN & RANSEN & RANDOC\\
\hline
ENG-RAW  & 4 & 26 & 17 & 7\\
ENG-LEMM  & 4 & 22 & 14 & 7\\
FRE-RAW  & 4 & 16 & 14 & 8\\
FRE-LEMM  & 4 & 14 & 12 & 8   \\
SPA-RAW  & 4 & 19 & 13 & 10\\
SPA-LEMM & 4 & 13 & 13 & 8\\
CHI & 4 & 10 & 6 & 5\\
\end{tabular}
\end{table*}

\begin{table*}[ht]
\caption{\label{tab:shortpath} Average Shortest Paths ($ \bar{d}$) for all networks. Standard deviation is shown in parenthesis}
\centering
\begin{tabular}{ccccc}
Statistic & COLL & PLAIN & RANSEN & RANDOC\\
\hline
ENG-RAW  & 2.48 (0.54) & 3.08 (0.72) & 2.95 (0.63) & 2.90 (0.56) \\
ENG-LEMM & 2.48 (0.54) & 3.02 (0.70) & 2.89 (0.60) & 2.83 (0.54) \\
FRE-RAW  & 2.49 (0.65) & 3.41 (0.83) & 3.07 (0.69) & 2.96 (0.58) \\
FRE-LEMM & 2.45 (0.58) & 3.30 (0.83) & 2.99 (0.69) & 2.86 (0.57) \\
SPA-RAW  & 2.50 (0.60) & 3.69 (0.98) & 3.06 (0.68) & 3.00 (0.62) \\
SPA-LEMM & 2.45 (0.59) & 3.19 (0.79) & 2.96 (0.68) & 2.88 (0.59) \\
CHI      & 2.30 (0.55) & 2.62 (0.71) & 2.32 (0.52) & 2.26 (0.54) \\
\end{tabular}
\end{table*}

\subsection{Clustering}

We examined two clustering coefficients for all networks 
The local clustering is defined as the mean of the vertex clustering,
\begin{equation}
C = \frac{1}{n}\sum_{i=1}^n C_i,
\label{clustering_equation}
\end{equation}
where $C_i$ is the clustering of the $i$-th vertex and $n$ is the number of vertices of the network. $C_i$ is the proportion of linkable pairs of adjacent vertices to vertex $i$ that are linked .

%I do not see compelling patterns. Further work needed due to tricky issues:
The global clustering coefficient is based on triplets of nodes. 
A triplet could be either open or closed. In an open triplet,
the three nodes are connected by two edges. Where as in a closed
triplet, they are connected with three edges. 
The global clustering coefficient is computing by calculating the ratio between the number of closed triplets and the total number of triplets. 
Both local and global clustering coefficients ignore the loops and the isolated vertices. View the networks as undirected.
The local and global clustering coefficients for all networks are shown in Tables \ref{tab:local_clustering} and \ref{tab:global_clustering} respectively.

%NEW: It you may want to compare the case of forbidden links and compare it with the case of loops allowed. When loops are forbidden you have to be careful: the local clustering should not include vertices with degree smaller than $2$. When loops are forbiden, you have to divide the sum of vertex clustering coefficients by the number of vertices with degree $2$ or greater, not by the raw number of vertices with degree one.  

\begin{table*}[ht]
\caption{\label{tab:global_clustering} Global Clustering Coefficients for all networks}
\centering
\begin{tabular}{ccccc}
Statistic & COLL & PLAIN & RANSEN & RANDOC\\
\hline
ENG & 0.0660 & 0.0236 & 0.0455 & 0.0492\\
STEMMED ENG & 0.0660 & 0.0548 & 0.1018 & 0.1086\\
FRE & 0.1465 & 0.0136 & 0.0341 &0.0429\\
STEMMED FRE & 0.1062 & 0.0273 & 0.0628 & 0.0764\\
SPA & 0.0730 & 0.0068 & 0.0193 & 0.0250\\
STEMMED SPA & 0.0500 & 0.0272 & 0.0521 & 0.0600\\
CHI & 0.2953 & 0.3509 & 0.2472 & 0.5043\\
\end{tabular}
\end{table*}

\begin{table*}[ht]
\caption{\label{tab:local_clustering} Local Clustering Coefficients for all networks}
\centering
\begin{tabular}{ccccc}
Statistic & COLL & PLAIN & RANSEN & RANDOC\\
\hline
ENG & 0.9605    & 0.5520 & 0.5910 & 0.6345\\
STEMMED ENG & 0.9605 & 0.5991 & 0.6582 & 0.7143\\
FRE & 0.7885 & 0.4550 & 0.5467 &0.5612\\
STEMMED FRE & 0.9235 & 0.4758 & 0.5745 & 0.6172\\
SPA & 0.7945 & 0.4017 & 0.5315 & 0.5094\\
STEMMED SPA & 0.8723 & 0.5044 & 0.5812 & 0.5844\\
CHI & 0.7688 & 0.5331 & 0.6101 & 0.8104\\
\end{tabular}
\end{table*}

\subsection{Degree correlations}

Assortative mixing is a bias in favor of connections between network nodes with similar characteristics. In other words the higher the degree of a vertex,
the higher  the degree of its neighbors. On the other hand, disassortative mixing refers to the phenomenon  where  the higher the degree of a vertex
the lower the degree of its neighbors.

\begin{figure}
\centering
\includegraphics[scale=0.40]{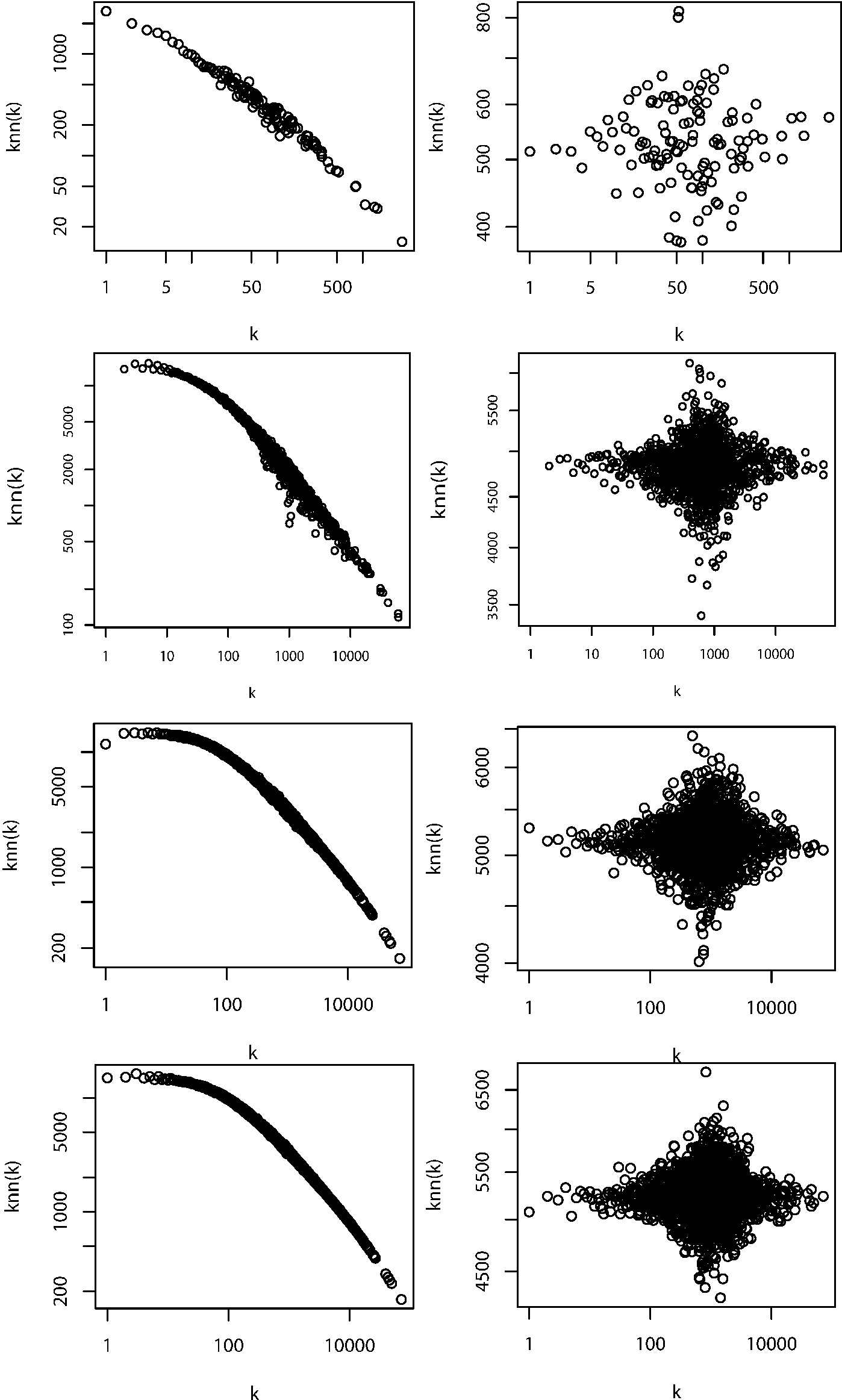}
\caption{\label{fig:eight}$\bar{k}_{nn}(k)$ for ENG-COLL, ENG, ENG-RANSEN, ENG-RANDOC and corresponding random networks}
\end{figure}

\begin{figure}
\centering
\includegraphics[scale=0.40]{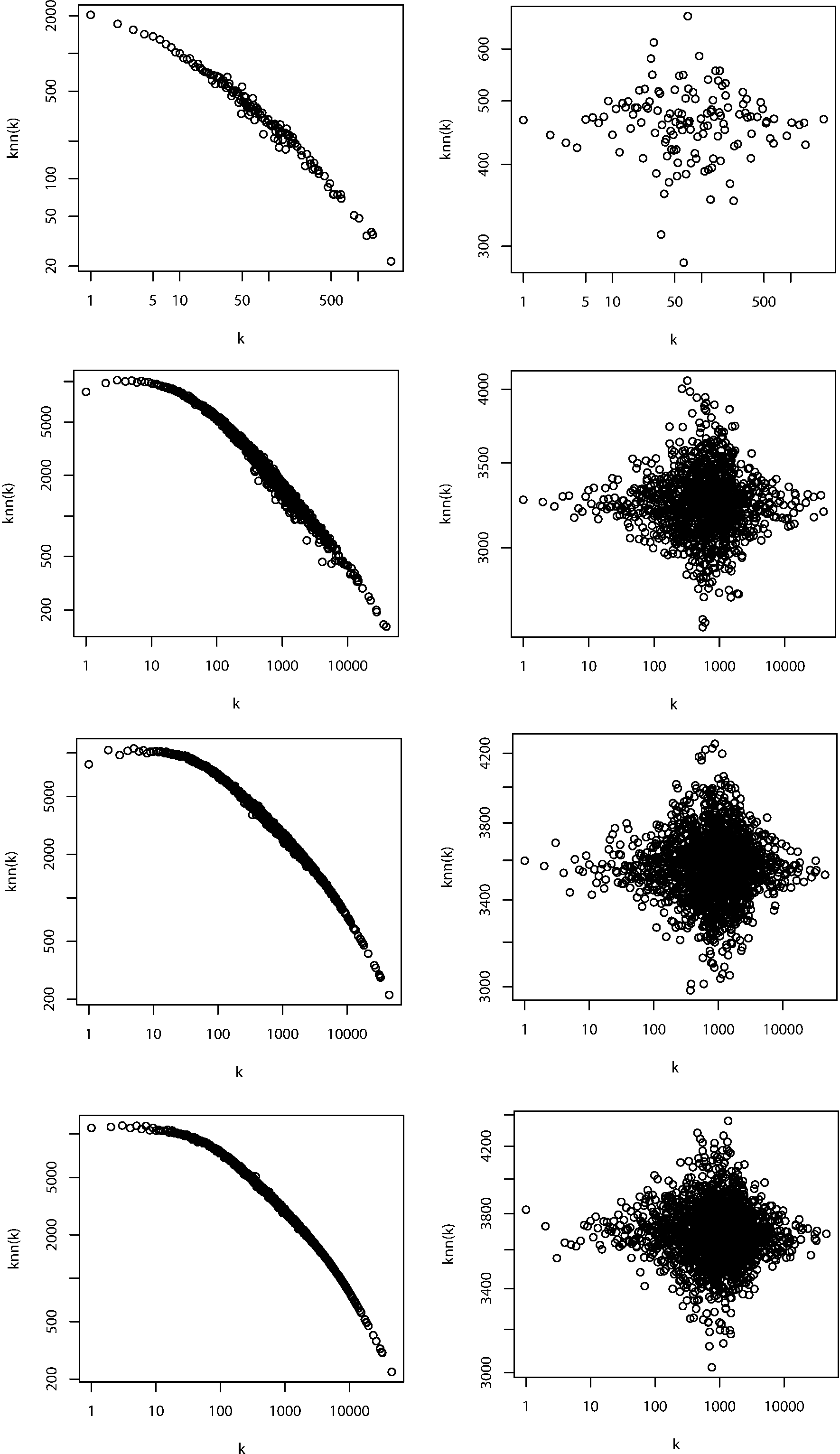}
\caption{\label{fig:fifteen}$\bar{k}_{nn}(k)$ for stemmed ENG-COLL, ENG, ENG-RANSEN, ENG-RANDOC and corresponding random networks}
\end{figure}

\begin{figure}
\centering
\includegraphics[scale=0.40]{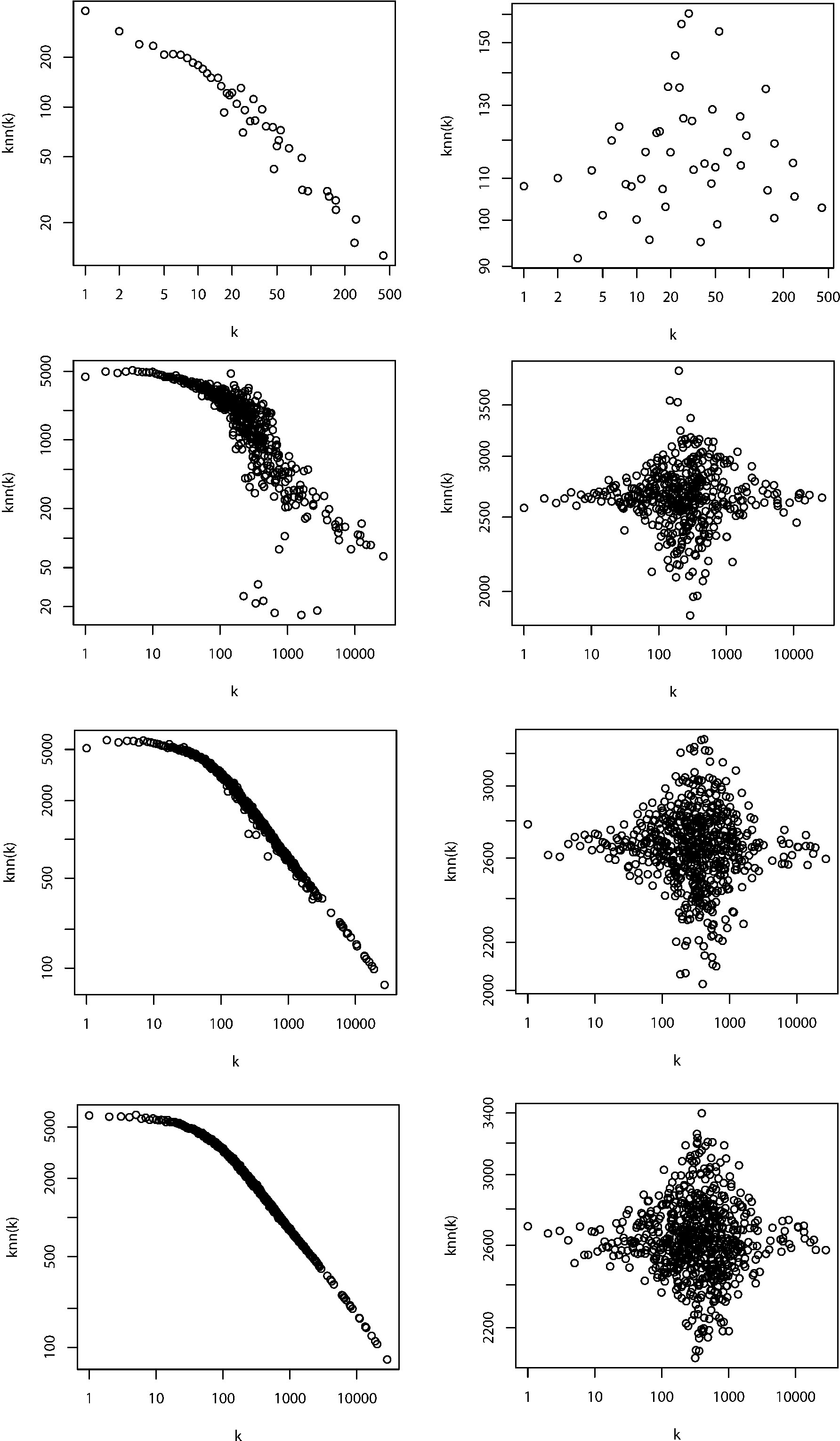}
\caption{\label{fig:sixteen}$\bar{k}_{nn}(k)$ for SPA-COLL, SPA, SPA-RANSEN, SPA-RANDOC and corresponding random networks}
\end{figure}

\begin{figure}
\centering
\includegraphics[scale=0.40]{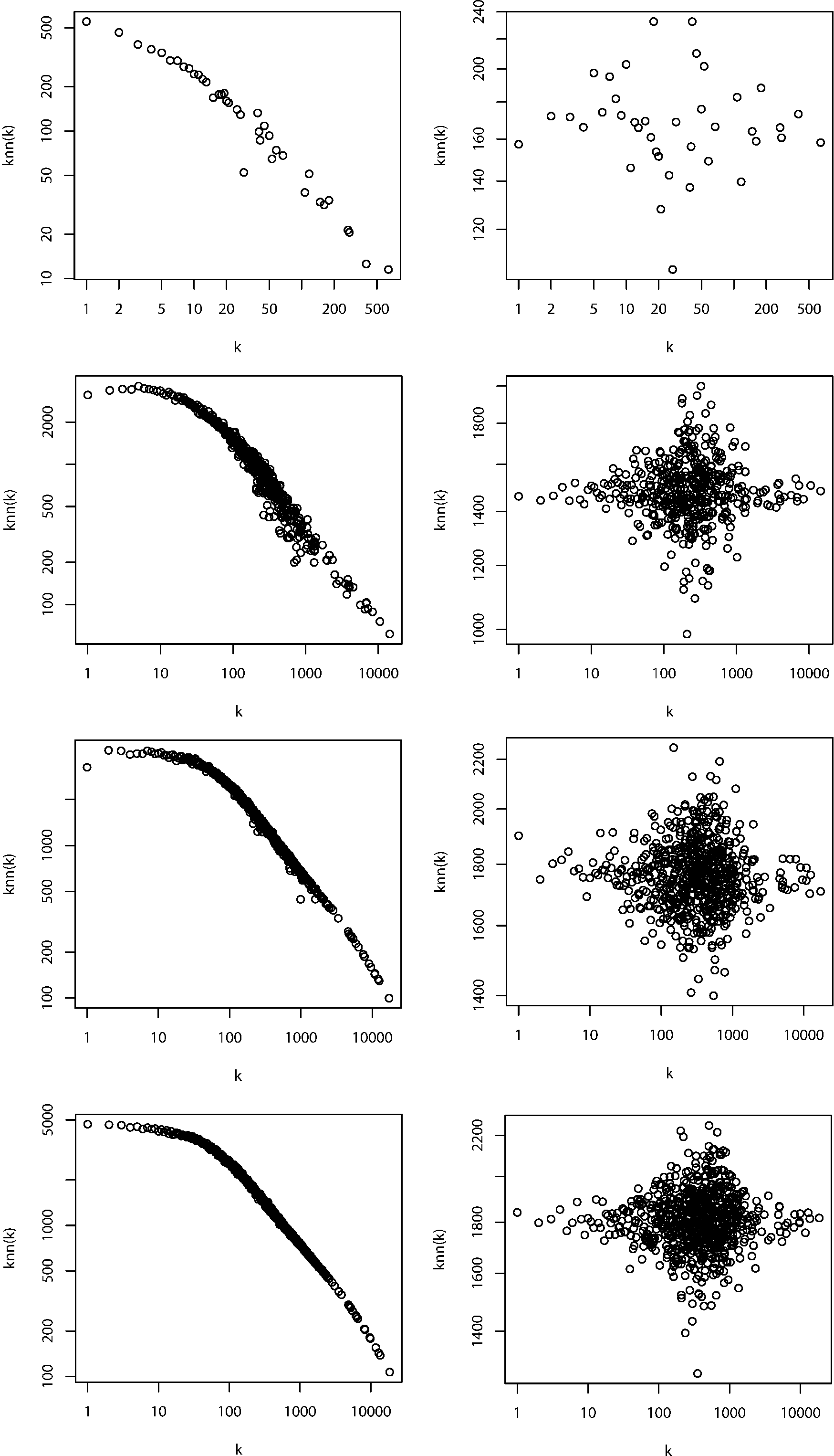}
\caption{\label{fig:seventeen}$\bar{k}_{nn}(k)$ for stemmed SPA-COLL, SPA, SPA-RANSEN, SPA-RANDOC and corresponding random networks}
\end{figure}

\begin{figure}
\centering
\includegraphics[scale=0.40]{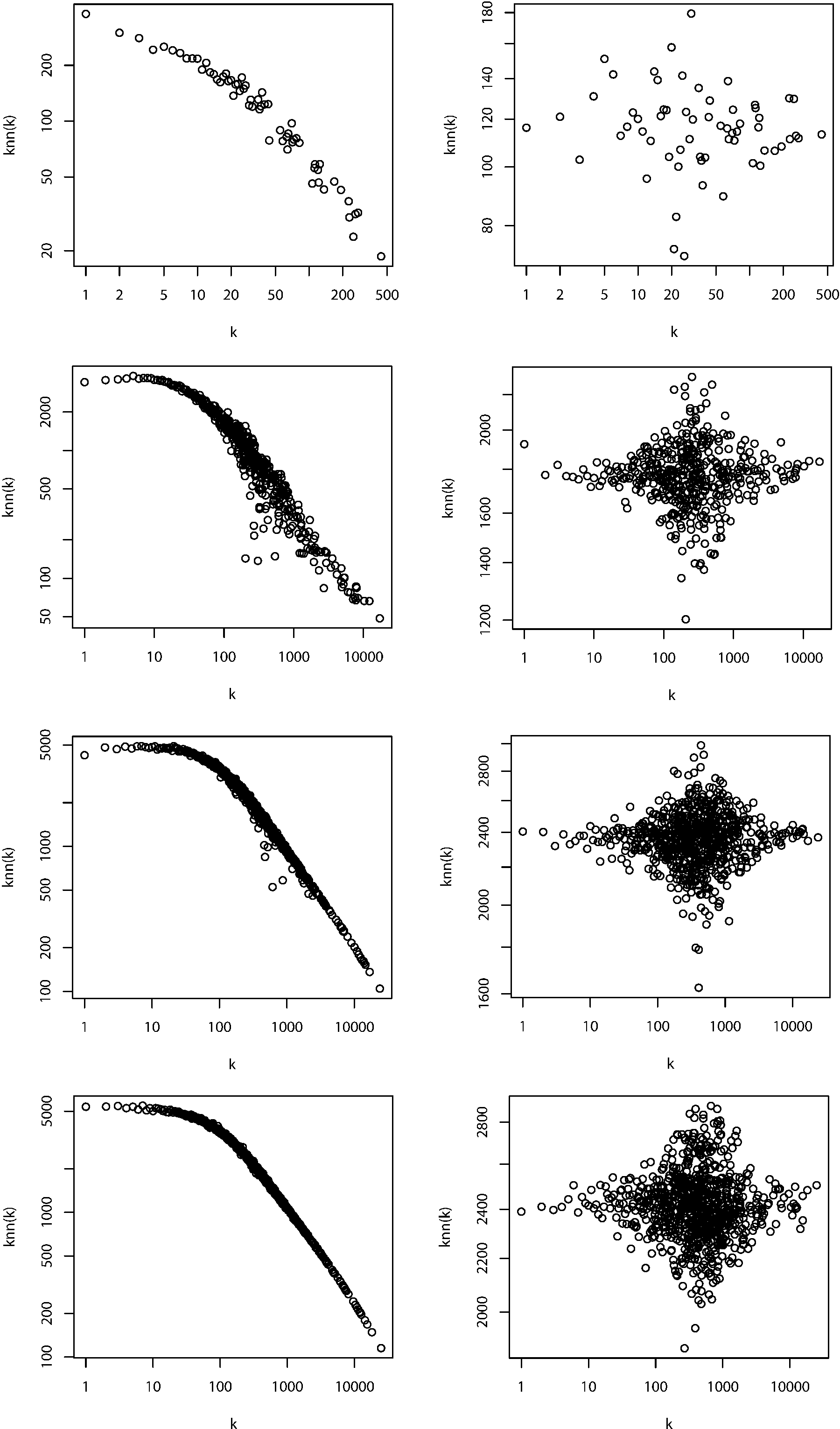}
\caption{\label{fig:eighteen}$\bar{k}_{nn}(k)$ for FRE-COLL, FRE, FRE-RANSEN, FRE-RANDOC and corresponding random networks}
\end{figure}

\begin{figure}
\centering
\includegraphics[scale=0.40]{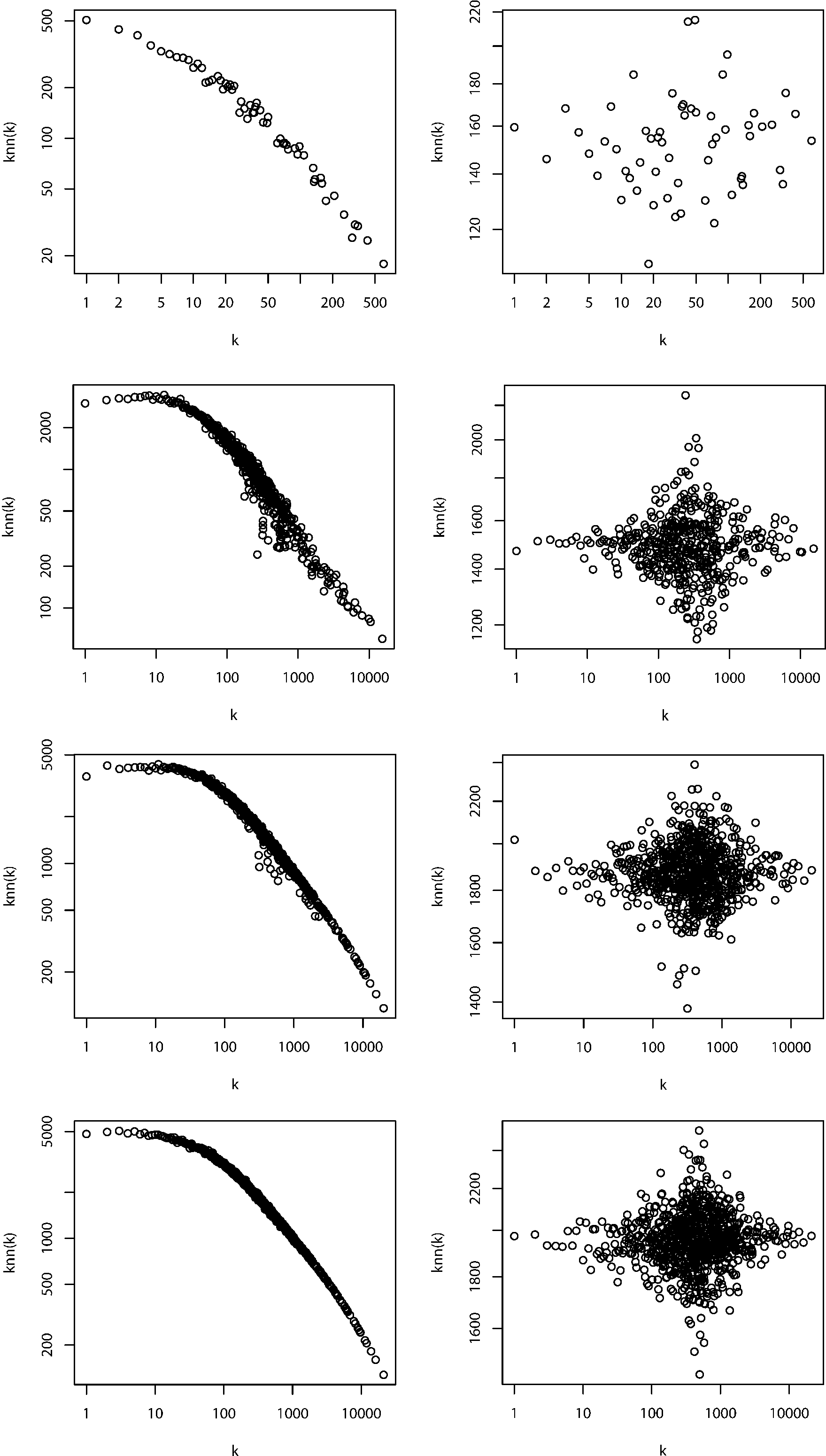}
\caption{\label{fig:nineteen}$\bar{k}_{nn}(k)$ for stemmed FRE-COLL, FRE, FRE-RANSEN, FRE-RANDOC and corresponding random networks}
\end{figure}

\begin{figure}
\centering
\includegraphics[scale=0.40]{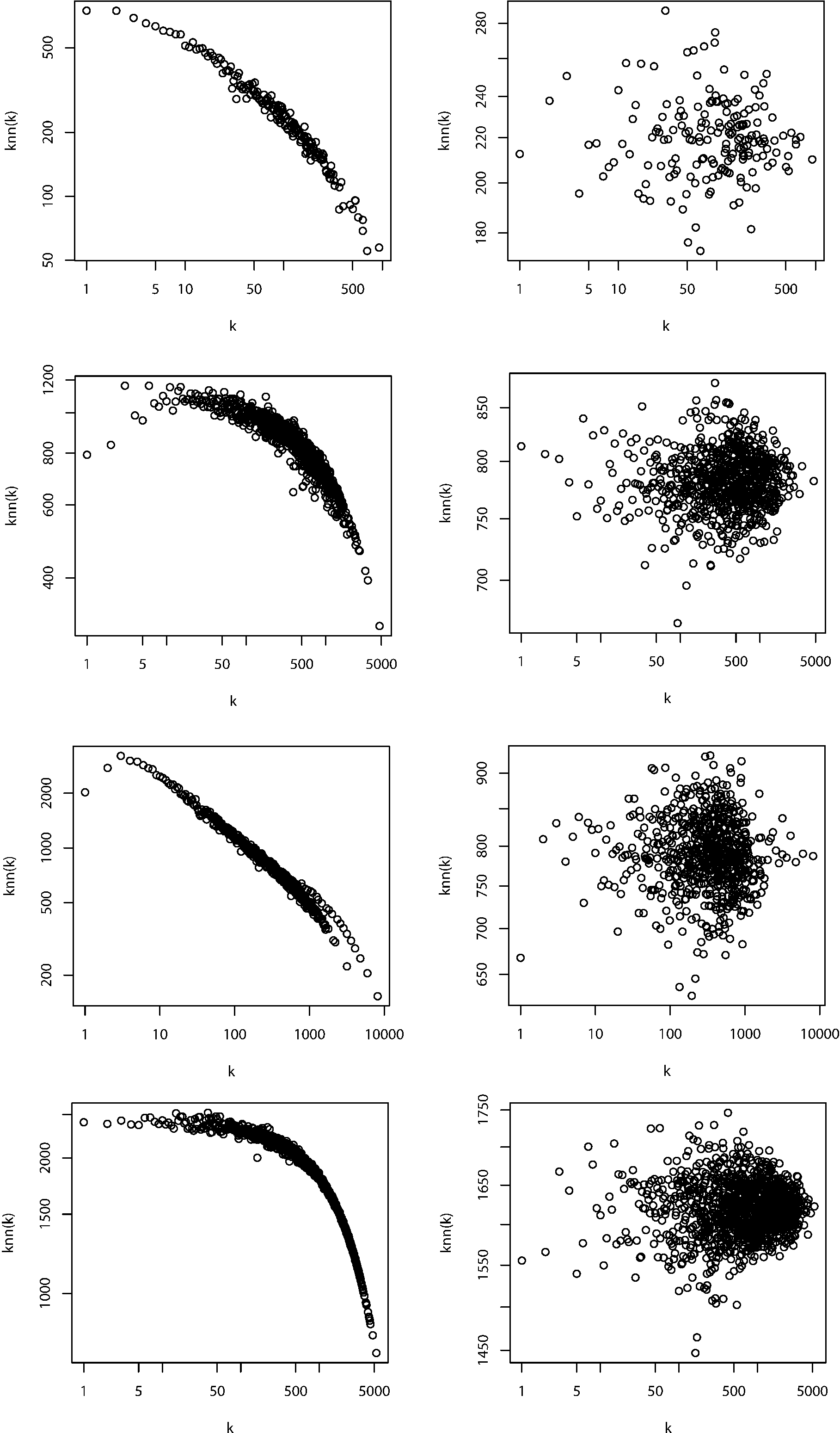}
\caption{\label{fig:twenty}$\bar{k}_{nn}(k)$ for CHI-COLL, CHI, CHI-RANSEN, CHI-RANDOC and corresponding random networks}
\end{figure}

Figures \ref{fig:eight},\ref{fig:fifteen},\ref{fig:sixteen},\ref{fig:seventeen},\ref{fig:eighteen},\ref{fig:nineteen} and \ref{fig:twenty}
show the relation between the degree of a vertex $k$ and the  normalized mean degree of the nearest neighbors of vertices of degree $k$.
We notice that the normalized mean degree of the nearest neighbors of vertices of degree k,
shrinks as $k$ grows for all networks, (i.e. vertices with large degree tend to connect with vertices with low degree).
The figures show that the networks exhibit disassortative mixing pattern for all languages and levels of constraint and regardless of the kind of the vertices (raw or lemmatized words).
The relationship between $k_{nn}$ and $k$ is a consequence of word frequencies. 
The network randomization algorithms destroys the pattern for all complexity levels.

%\item
%Is the fact that loops are forbidden contributing to some strange effect? (see below).
%\item
%Dragomir: any better idea?    
%\end{itemize}

%Tricky issues: Loops (a link from a vertex to itself). Loops should not be removed to not force correlations (this can be problematic at the level of clustering, $k_{nn}$,... Dragomir: do you remove loops? Does the network randomization algorithm allow a vertex to become connected with itself when it was not? Is the algorithm able to work with networks with loops?

\subsection{The relationship between vertex frequency and degree}

Please, put a table showing the 5 or 10 most frequent words in the English corpus and the 5 or 10 most connected words in each of the networks: COLL, PLAIN, RANSEN and RANDOC. 

\begin{figure}
\centering
\includegraphics[scale=0.60]{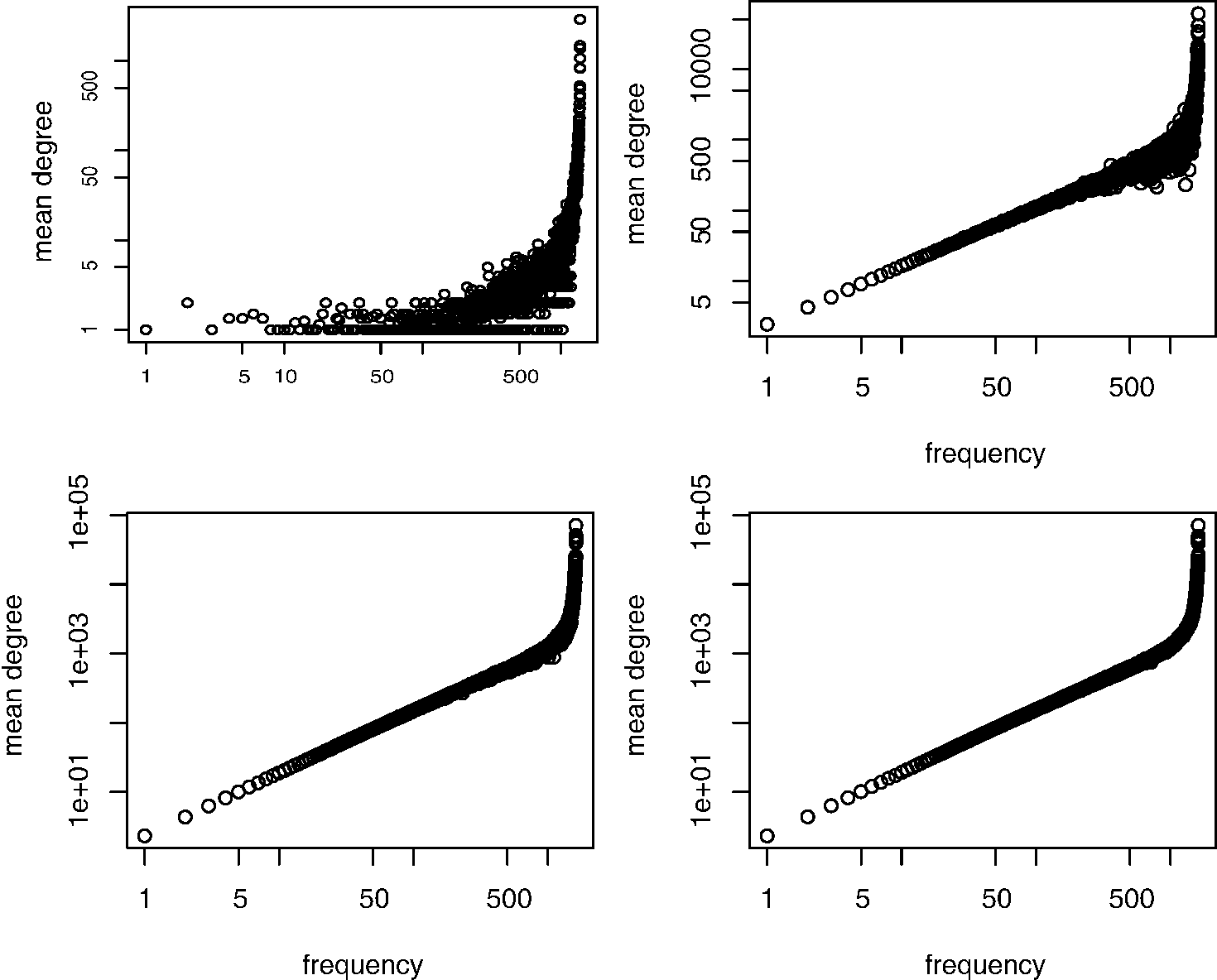}
\caption{\label{fig:five}Averaged Degree by word frequency for ENG}
\end{figure}

\begin{figure}
\centering
\includegraphics[scale=0.30]{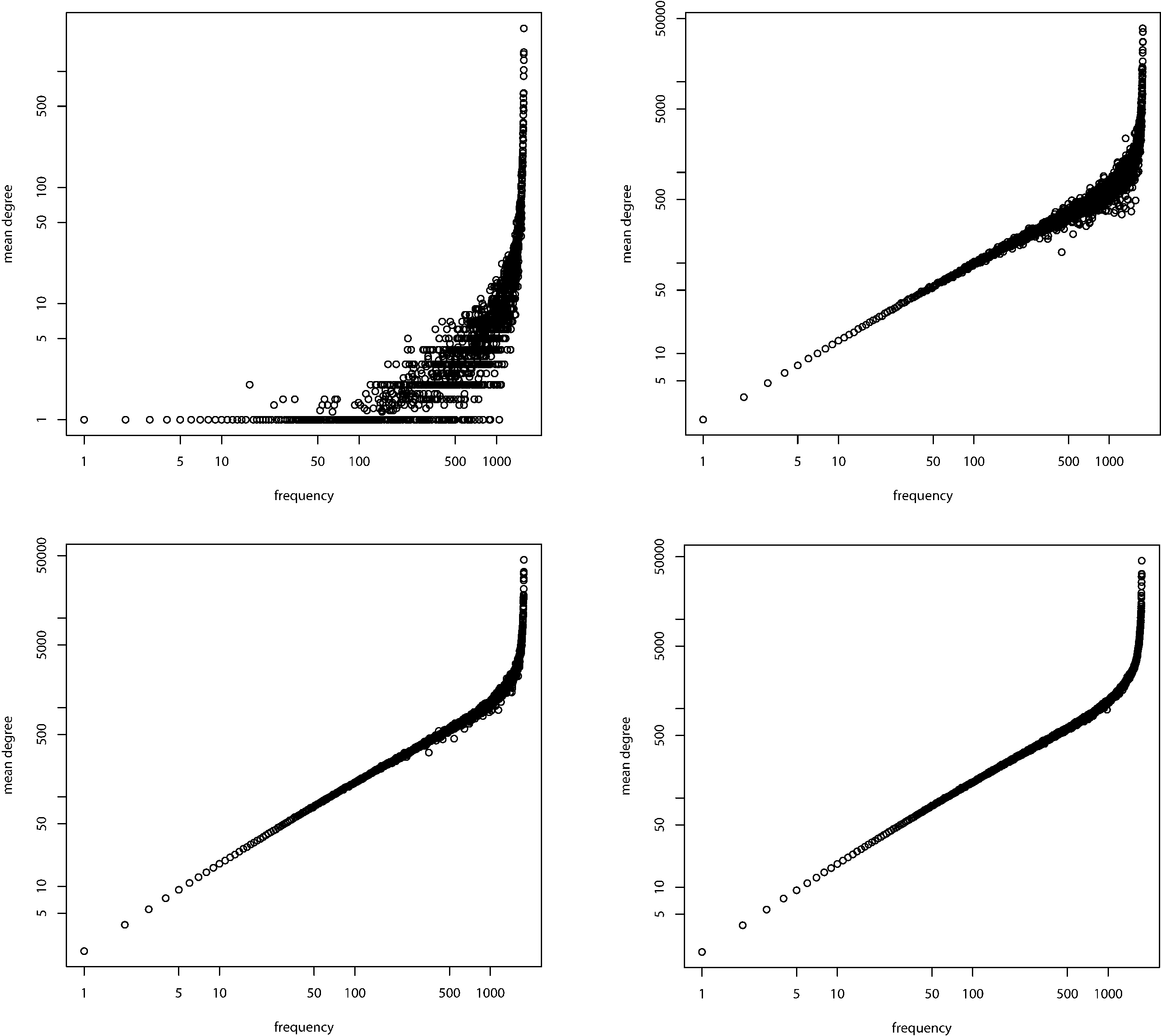}
\caption{\label{fig:five2}Averaged Degree by word frequency for stemmed ENG}
\end{figure}

\begin{figure}
\centering
\includegraphics[scale=0.30]{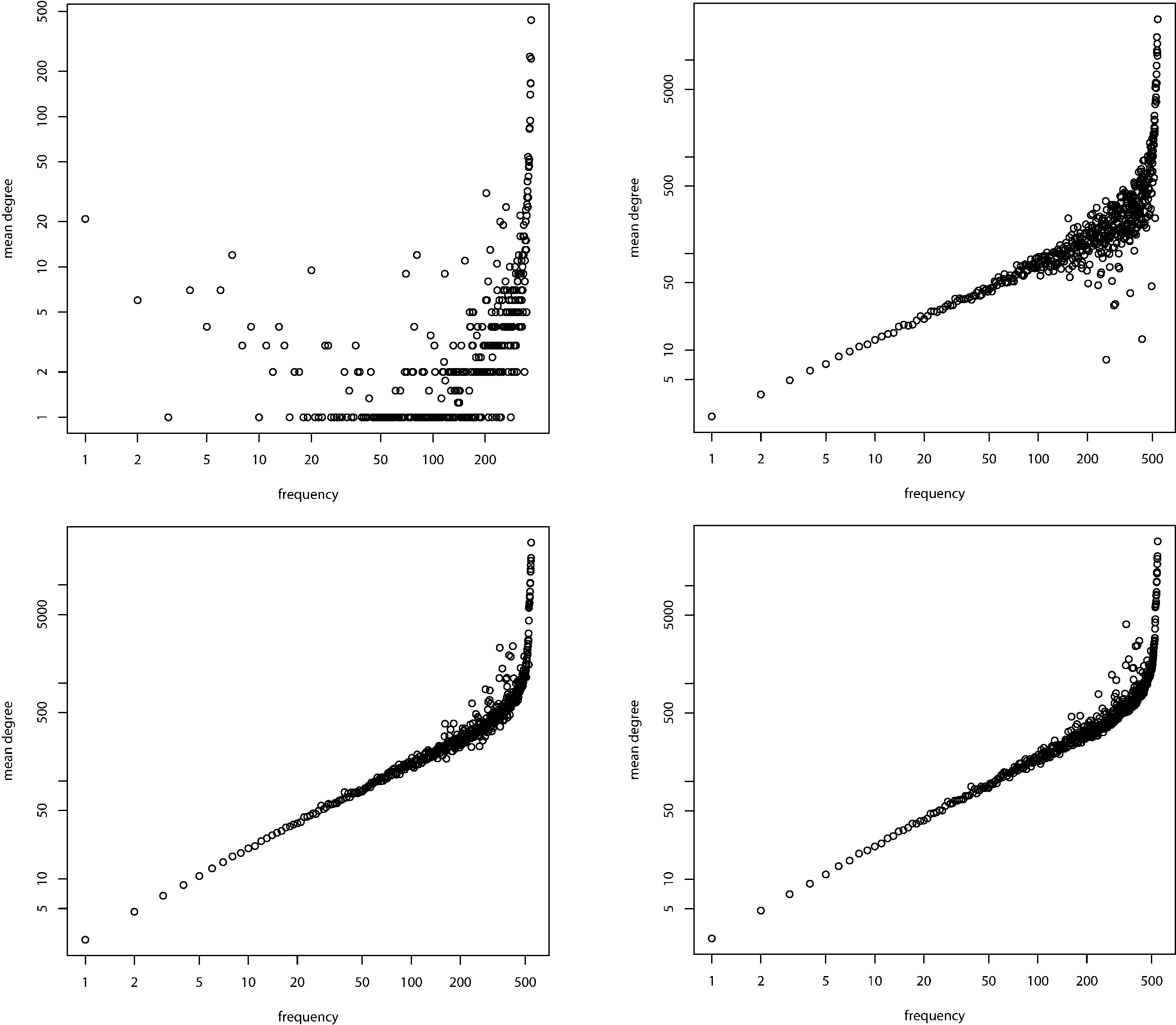}
\caption{\label{fig:five3}Averaged Degree by word frequency for SPA}
\end{figure}

\begin{figure}
\centering
\includegraphics[scale=0.30]{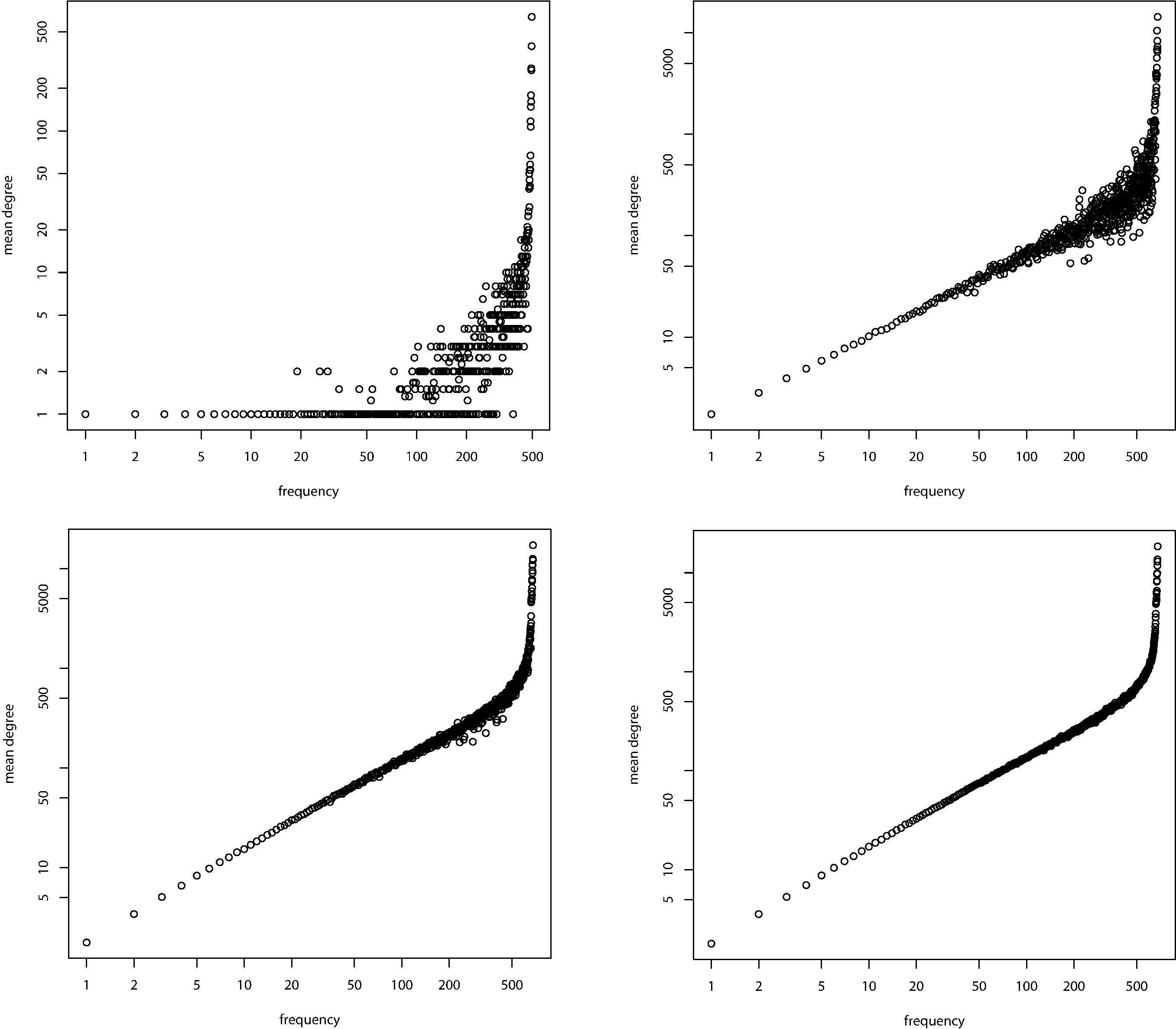}
\caption{\label{fig:five4}Averaged Degree by word frequency for stemmed SPA}
\end{figure}

\begin{figure}
\centering
\includegraphics[scale=0.30]{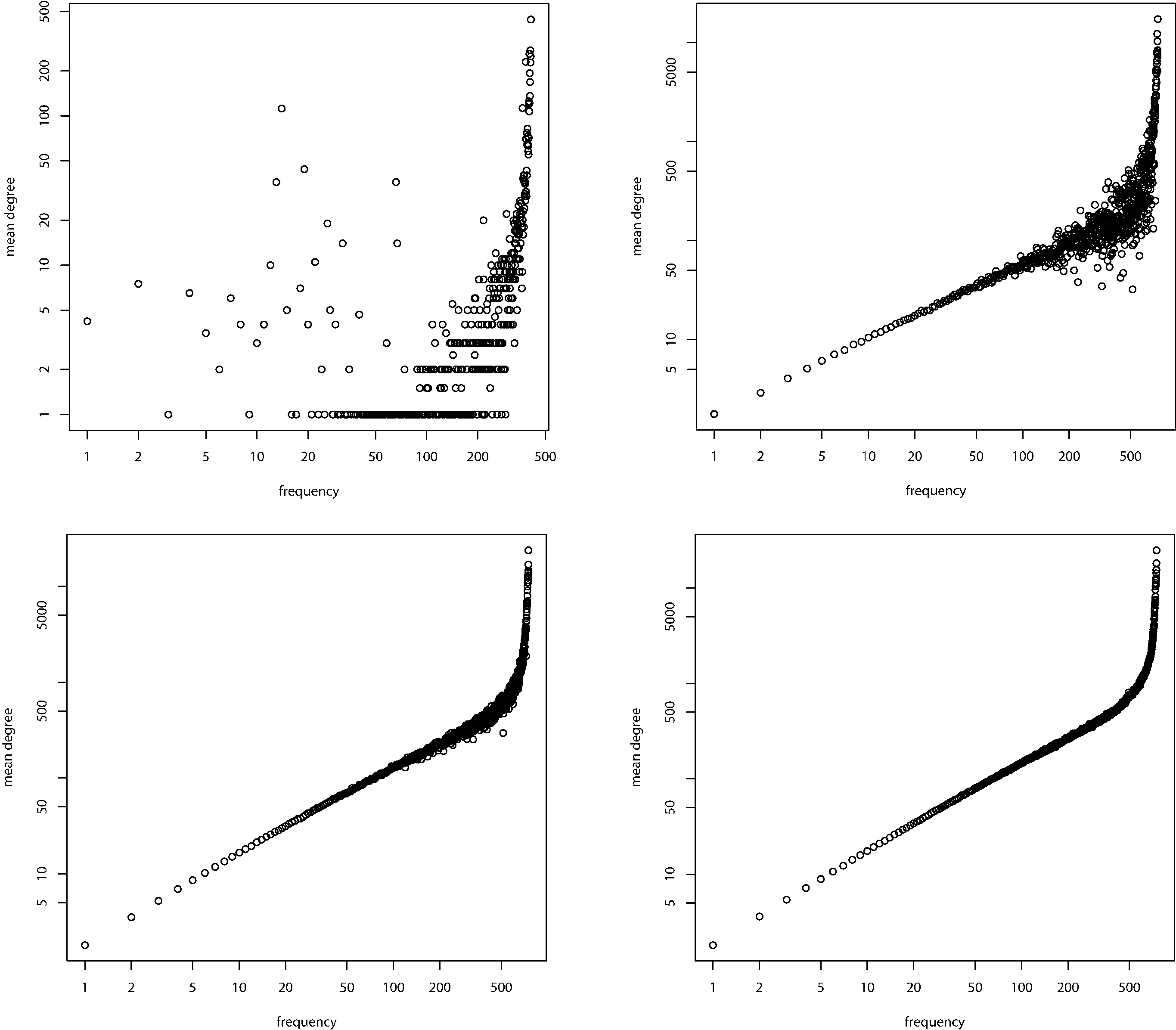}
\caption{\label{fig:five5}Averaged Degree by word frequency for FRE}
\end{figure}

\begin{figure}
\centering
\includegraphics[scale=0.30]{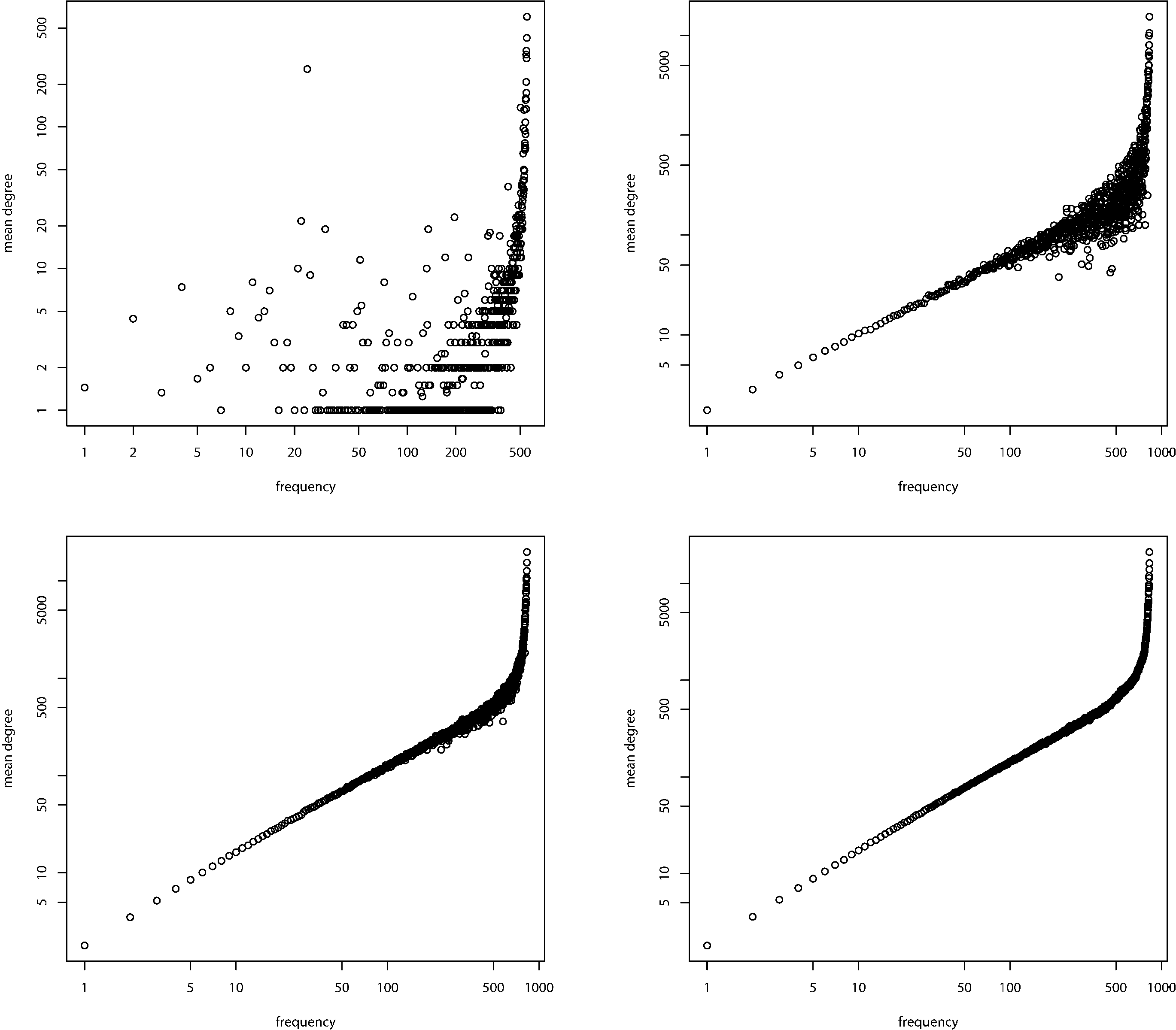}
\caption{\label{fig:five6}Averaged Degree by word frequency for stemmed FRE}
\end{figure}

\begin{figure}
\centering
\includegraphics[scale=0.30]{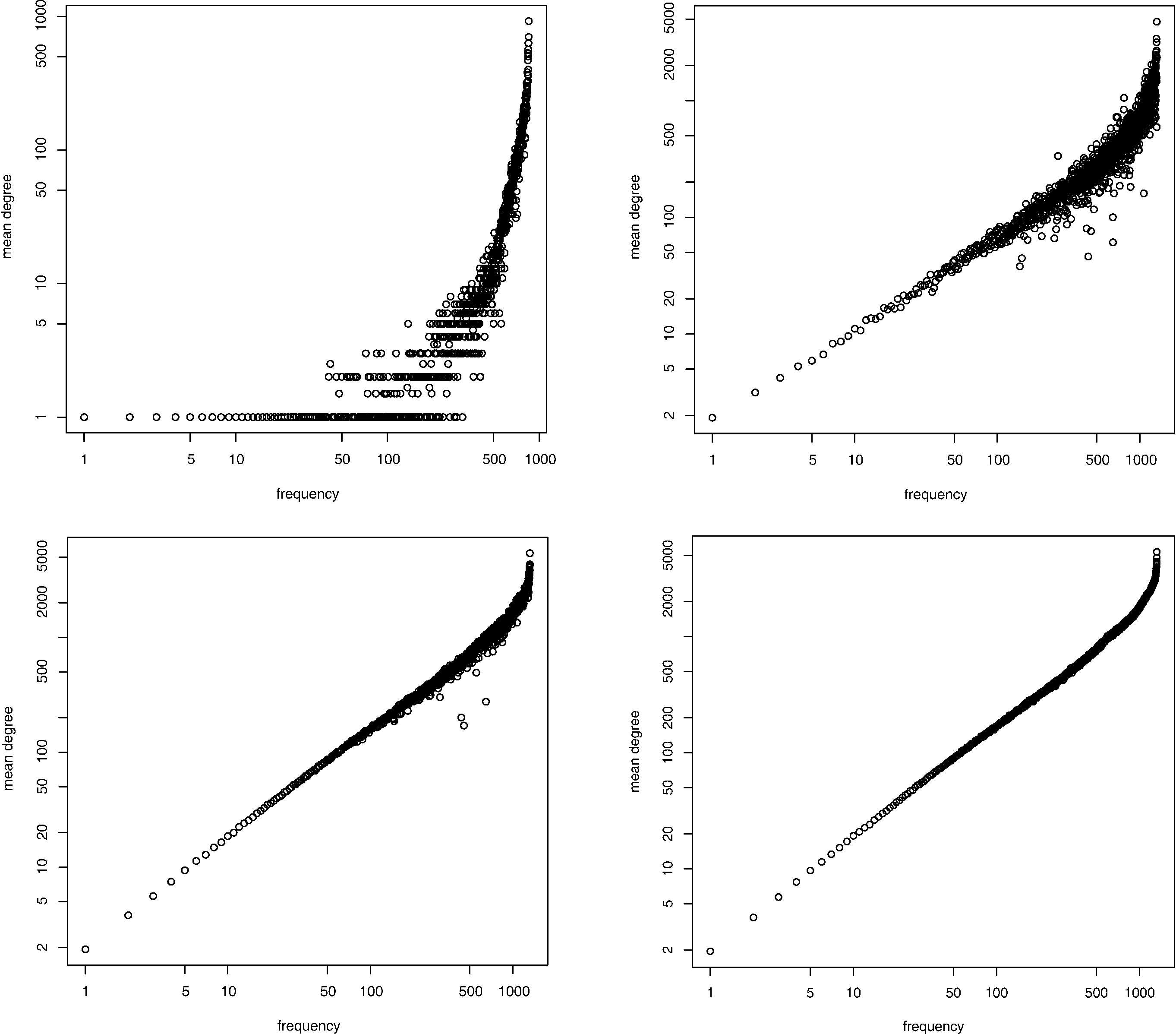}
\caption{\label{fig:five7}Averaged Degree by word frequency for CHI}
\end{figure}

To examine the relationship between vertex frequency and degree, we use linear regression after excluding
some outlier with $frequency > 10^5$.
The result of linear regression
for the real network shows a strong correlation between degree and frequency. The slope is 2.86, the $p$ value is less than $2*10^{-16}$, $R^2$ is 0.89.
For the random network, we also observe a high correlation between degree and frequency. The slope is 2.11, the p value of
the correlation between two variables is less than $2*10^{-16}$, $R^2$ is 0.89.
In addition, we averaged the degree of words of a certain frequency and then make a plot for all the frequency range for
all networks(Fig. \ref{fig:five}, Fig. \ref{fig:five2}, Fig. \ref{fig:five3}, Fig. \ref{fig:five4}, Fig. \ref{fig:five5}, Fig. \ref{fig:five6}, Fig. \ref{fig:five7}).

We notice that the  higher the frequency of a word, the higher its degree. We also notice that 
the hubs of the networks are high frequency words. 
Lists of the top ten frequent words and the top ten most connected words in all English networks are
shown in Tables \ref{tab:fre}, \ref{tab:colltopten}, \ref{tab:engtopten}, \ref{tab:ransentopten}, and \ref{tab:randoctopten}.

\section{Conclusion}

We studied the topological properties of linguistics networks
at different levels of linguistic constraints.
We found out that the networks produced
from randomized data exihibit small worlds and  scale-free
characteristics. One possible explanation would be that 
degree distributions are functions of the word frequencies.
However human language is a very complex ``system'' and there is no simple way
to explain this observation. 
We also find out hat the degree
distribution of a co-occurrence graph does change under randomization
Further, network
statistics such as diameter and clustering coefficient too seem to
depend on the degree distributions of the underlying network.

\section*{Acknowledgements}
The authors would like to thank Ramon Ferrer i Cancho for his valuable comments and contributions.

\bibliography{nets}

\end{document}